\documentclass[letterpaper]{article} 
\usepackage{aaai2026}
\usepackage{times}  
\usepackage{helvet}  
\usepackage{courier}  
\usepackage[hyphens]{url}  
\usepackage{graphicx} 
\usepackage{natbib}  
\usepackage{caption} 
\usepackage{algorithm}
\usepackage{algorithmic}

\usepackage{multirow}
\usepackage{booktabs}

\usepackage{amsmath}
\usepackage{amssymb}
\usepackage{diagbox}

\usepackage{newfloat}
\usepackage{listings}
\DeclareCaptionStyle{ruled}{labelfont=normalfont,labelsep=colon,strut=off} 
\floatstyle{ruled}
\newfloat{listing}{tb}{lst}{}
\floatname{listing}{Listing}
\nocopyright 

\title{GeM-VG: Towards Generalized Multi-image Visual Grounding with Multimodal Large Language Models}
\author {
    Shurong Zheng\textsuperscript{\rm 1, \rm 2 \rm 3},
    Yousong Zhu\textsuperscript{\rm 4},
    Hongyin Zhao\textsuperscript{\rm 1},
    Fan Yang\textsuperscript{\rm 1, \rm 2 \rm 3},
    Yufei Zhan\textsuperscript{\rm 1, \rm 3},
    Ming Tang\textsuperscript{\rm 1},
    Jinqiao Wang\textsuperscript{\rm 1, \rm 2, \rm 3}\thanks{Corresponding author.},
}
\affiliations {
    \textsuperscript{\rm 1} Foundation Model Research Center, Institute of Automation, Chinese Academy of Sciences, Beijing, China\\
    \textsuperscript{\rm 2}Peng Cheng Laboratory, Shenzhen, China\\
    \textsuperscript{\rm 3}School of Artificial Intelligence, University of Chinese Academy of Sciences, Beijing, China\\
    \textsuperscript{\rm 4}School of Artificial Intelligence, China University of Mining and Technology-Beijing, Beijing, China\\
    \textsuperscript{\rm 5}Wuhan AI Research, Wuhan, China\\
    zhengshurong2023@ia.ac.cn, yousong.zhu@cumtb.edu.cn,  zhaohongyin2020@ia.ac.cn, yangfan\_2022@ia.ac.cn, zhanyufei2021@ia.ac.cn, tangm@nlpr.ia.ac.cn, jqwang@nlpr.ia.ac.cn
}
\usepackage{bibentry}

\begin{document}

\maketitle
\begin{abstract}
Multimodal Large Language Models (MLLMs) have demonstrated impressive progress in single-image grounding and general multi-image understanding. Recently, some methods begin to address multi-image grounding. However, they are constrained by single-target localization and limited types of practical tasks, due to the lack of unified modeling for generalized grounding tasks.
Therefore, we propose GeM-VG, an MLLM capable of \textbf{Ge}neralized \textbf{M}ulti-image \textbf{V}isual \textbf{G}rounding.
To support this, we systematically categorize and organize existing multi-image grounding tasks according to their reliance of cross-image cues and reasoning, and introduce the MG-Data-240K dataset, addressing the limitations of existing datasets regarding target quantity and image relation.
To tackle the challenges of robustly handling diverse multi-image grounding tasks, we further propose a hybrid reinforcement finetuning strategy that integrates chain-of-thought (CoT) reasoning and direct answering, considering their complementary strengths.
This strategy adopts an R1-like algorithm guided by a carefully designed rule-based reward, effectively enhancing the model’s overall perception and reasoning capabilities.
Extensive experiments demonstrate the superior generalized grounding capabilities of our model. For multi-image grounding, it outperforms the previous leading MLLMs by 2.0\% and 9.7\% on MIG-Bench and MC-Bench, respectively. In single-image grounding, it achieves a 9.1\% improvement over the base model on ODINW. Furthermore, our model retains strong capabilities in general multi-image understanding.

\end{abstract}    
\section{Introduction}

\begin{figure}[ht!]
\centering
\includegraphics[width=1.0\linewidth]{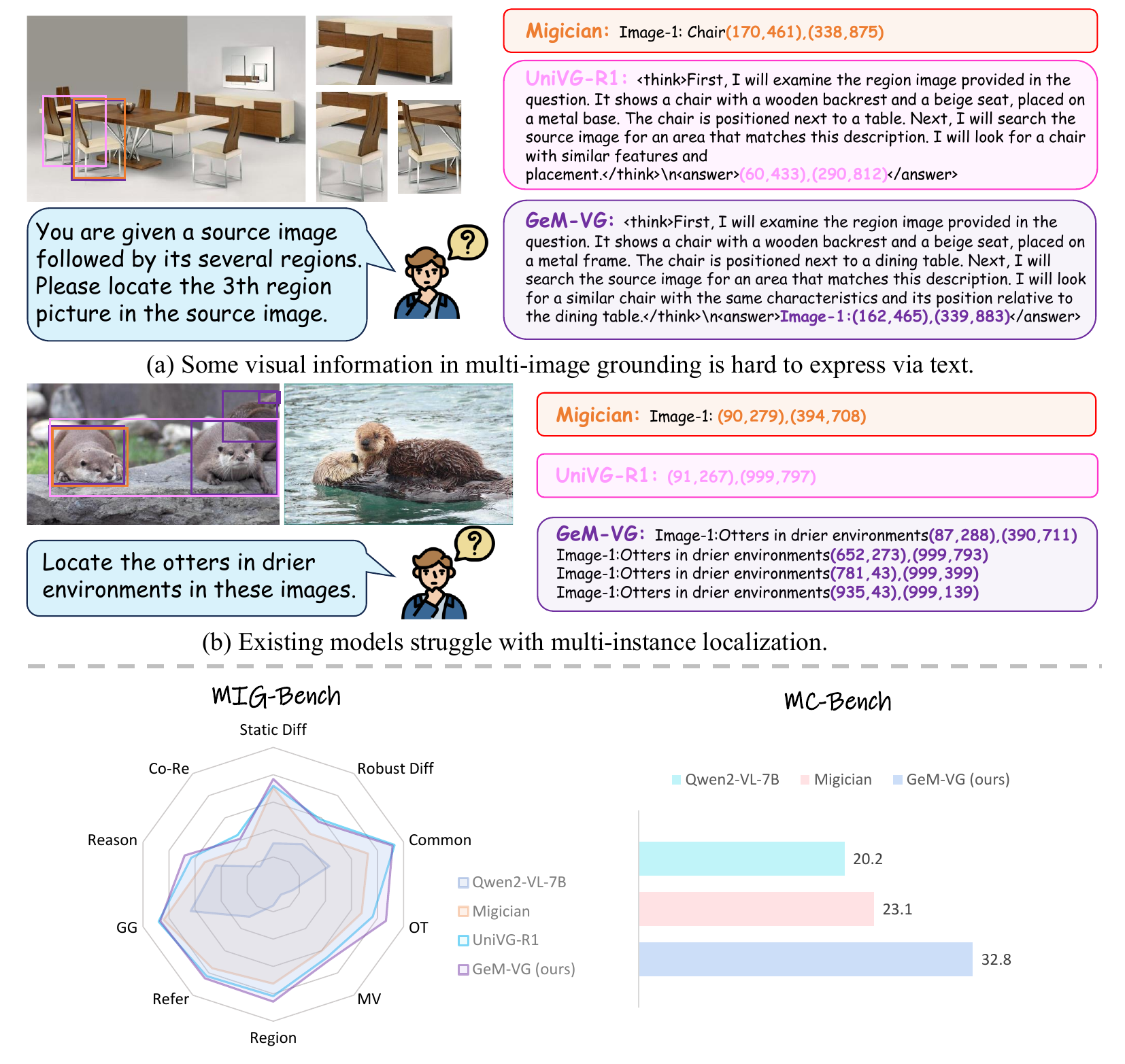}
\caption{\textbf{(Top)} The limitations of existing multi-image grounding models: (a) The reasoning-guided model struggles with fine-grained perception tasks, where the visual cues are hard to verbalize. (b) They fail to localize all referring instances in multi-object scenarios. \textbf{(Bottom)} Our model, GeM-VG, achieves strong performance across a range of multi-image grounding tasks.}
\label{fig:head}
\end{figure}

Traditional visual grounding encompasses tasks such as referring expression comprehension~\cite{yu2016refcoco, nagaraja2016refcocog}, phrase grounding~\cite{plummer2015flickr30k}, and object detection~\cite{lin2014mscoco}, which involve localizing target regions within a single image based on simple textual descriptions. 
With the advancements in MLLMs, some works~\cite{chen2023shikra, zhan2024griffon, peng2023kosmos-2, you2023ferret, zhan2024griffonv2} leverage the powerful multimodal comprehension capabilities of MLLMs to facilitate visual grounding tasks. Despite their effectiveness, these works are limited to single-image scenarios.

The capabilities of multi-image perception and reasoning are essential for real-world applications such as autonomous driving and GUI agents, which require analyzing contextual information extracted from multiple images. While some works~\cite{li2024llava-next-interleave, jiang2024mantis, li2024llava-onevision, ye2024mplug-owl3} investigate image-level tasks in multi-image scenarios, region-level comprehension remains relatively underexplored. 
Recently, Migician~\cite{li2025migician} introduces a benchmark comprising diverse multi-image grounding tasks, laying a foundation for this emerging direction. UniVG-R1~\cite{bai2025univg-r1} further enhances reasoning via reinforcement learning (RL). However, these works focus predominantly on single-target grounding. In practical scenarios, the number of referring objects can vary (e.g., "the otters in drier environments" in Figure~\ref{fig:head}(b)).

In this work, we aim to advance generalized visual grounding in multi-image scenarios, and propose GeM-VG, an MLLM for \textbf{Ge}neralized \textbf{M}ulti-image \textbf{V}isual \textbf{G}rounding.
To cover comprehensive multi-image grounding scenarios, we systematically categorize existing multi-image grounding tasks based on cognitive requirements and image relations. Due to the limited target quantity and task scenarios in previous multi-image grounding datasets~\cite{li2025migician, bai2025univg-r1}, we curate a multi-image grounding dataset MG-Data-240K to support broader scenarios. 
As multi-image grounding involves increasingly complex multimodal contexts, it requires more sophisticated perception and reasoning. Inspired by the recent success of reasoning models~\cite{jaech2024openai-o1, guo2025deepseek-r1}, we design a new rule-based reward and adapt the R1-like reinforcement learning method to generalized grounding tasks.
Due to the lack of basic multi-image grounding capability in the base model, 
we adopt a progressive three-stage training strategy: (1) supervised finetuning with short-answer data to build fundamental grounding capabilities, (2) CoT-based supervised finetuning to guide reasoning, and (3) GRPO training guided by multi-dimensional visual feedback to further enhance localization and reasoning.

Through supervised fine-tuning with data in different answering modes, we find that CoT reasoning does not consistently outperform direct-answer. In tasks relying on detailed perceptual cues, directly predicting locations avoids the ambiguity of verbalizing abstract visual cues (e.g., Figure~\ref{fig:head}). Moreover, CoT provides limited benefits in tasks with minimal reasoning demands. To leverage the complementary strengths of both modes, we propose a hybrid finetuning strategy that encourages exploration and optimization of both answering modes during training.

We evaluate GeM-VG on two multi-image grounding benchmarks: MIG-Bench~\cite{li2025migician} and MC-Bench~\cite{xu2024mc-bench}. On MIG-Bench, GeM-VG surpasses the previous state-of-the-art by 2.0\%. On MC-Bench, it outperforms Migician by 9.7\% and the base model Qwen2VL-7B by 12.6\%. Additionally, our model achieves consistent improvements on single-image grounding and general multimodal understanding tasks.
The main contributions of this work are summarized as follows:
\begin{enumerate}
\item We provide a systematic taxonomy of existing multi-image grounding tasks and introduce the MG-Data-240K dataset, which encompasses diverse tasks and varying numbers of targets.

\item We propose GeM-VG, an MLLM for generalized multi-image grounding, along with a hybrid RL finetuning strategy that integrates both chain-of-thought and direct-answer modes, tailored for grounding tasks with arbitrary numbers of targets.

\item Extensive experimental results demonstrate that our method consistently improves the performance across various multi-image grounding tasks, while maintaining strong generalization capabilities.
\end{enumerate}

\begin{figure*}
	\centering
	\begin{minipage}[t!]{\linewidth}
		\centering
		\includegraphics[width=1.0\linewidth]{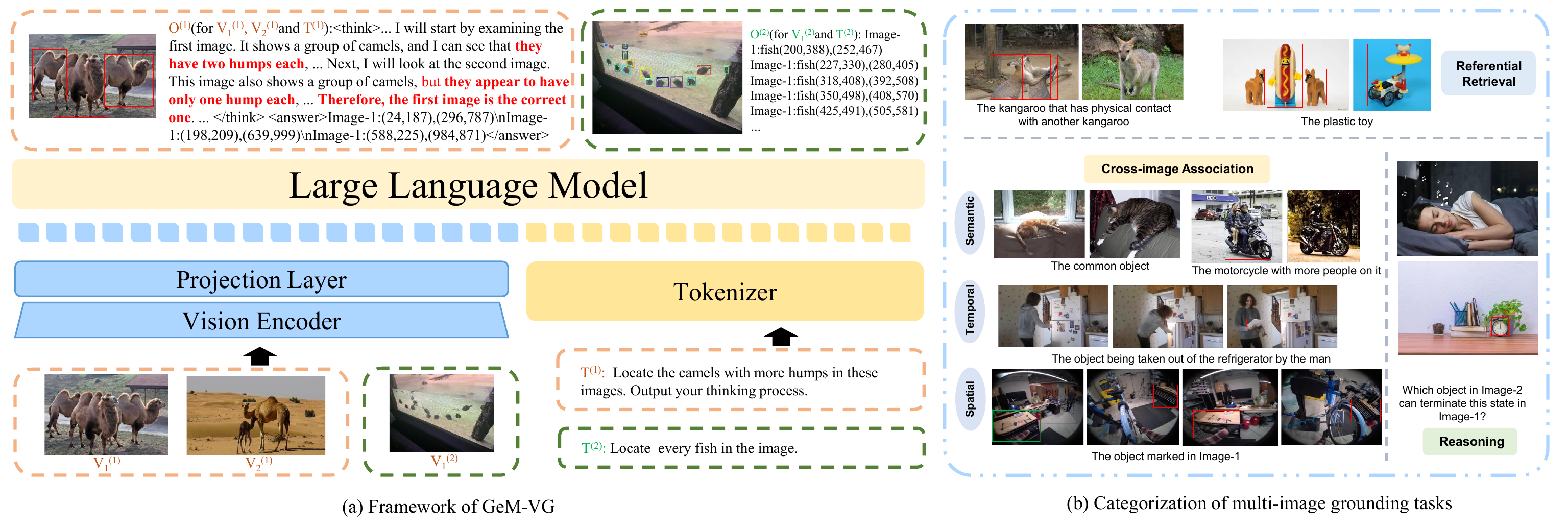}
		\caption{\textbf{(Left)} Overview of GeM-VG. GeM-VG is capable of reasoning over multiple visual contexts and localizing all referring instances. \textbf{(Right)} We provide a systematic taxonomy of generalized multi-image grounding scenarios.} 
        \label{fig:model}
	\end{minipage}
\end{figure*}

\section{Related Work}
\subsection{Multimodal Large Language Models for Visual Grounding}
Recent advances in MLLMs have led to impressive progress in general vision-language understanding and reasoning. By leveraging the powerful multimodal comprehension capabilities of MLLMs, visual grounding tasks are also innovated. Some works~\cite{chen2023shikra, zhan2024griffon, you2023ferret} focus on enabling MLLMs to support a range of grounding tasks. With improved reasoning capacity, some studies~\cite{liu2025visual-rft, ma2025deepperception} explore complex reasoning grounding. However, these works are limited to single-image scenarios, hindering the applicability in broader real-world scenarios. Recently, Migician~\cite{li2025migician} introduces a multi-image grounding benchmark and enables the model to perform free-form multi-image grounding via supervised finetuning. UniVG-R1~\cite{bai2025univg-r1} further enhances reasoning via reinforcement learning. Nevertheless, they primarily focus on single-object grounding and struggle with multi-instance cases. In contrast, our work aims to address generalized multi-image grounding tasks.

\subsection{Vision-Language Reinforcement Learning}
With the emergence of large reasoning models, reinforcement learning has become a research focus for enhancing the reasoning capabilities of LLMs. Recently, DeepSeek-R1~\cite{guo2025deepseek-r1} achieves a breakthrough in this area by introducing a new rule-based RL algorithm.
Inspired by the success of DeepSeek-R1, a series of works apply the rule-based GRPO method to vision-language domain. Among these, some focus on visual reasoning tasks such as mathematical reasoning~\cite{huang2025vision-r1, deng2025openvlthinker, wang2025vl-rethinker, zhang2025r1-vl, yang2025r1-onevision}. Some other studies investigate visual perception tasks including visual grounding~\cite{shen2025vlm-r1, zhan2025vision-r1, ma2025deepperception, liu2025visual-rft, yu2025perception-r1}, which are closely related to our work. However, they are limited to single-image scenarios, lacking the ability to perform precise grounding across multiple images. In this work, we explore the R1-like paradigm's potential in multi-image grounding.
\section{Methodology}


\subsection{Overview}
\label{sub: overview}
We start with a formal definition of the generalized multi-image grounding task and a brief overview of our GeM-VG. Formally, given a sequence of images $V=\{V_1, V_2, ..., V_m\}$ and a textual instruction $T$, the model $\mathcal{M}$ is required to localize all relevant instances. The output is a set of bounding boxes and their corresponding image indices:
\begin{equation}
O = \{(b_1, i_1), (b_2, i_2), ..., (b_n, i_n)\} = \mathcal{M}(V,T),
\end{equation}

\noindent where $b_k$ denotes the $k$-th bounding box and $i_k \in \{1, ..., m\}$ indicates the index of the corresponding image $V_{i_k}$. 

As depicted in Figure~\ref{fig:model}(a), we develop GeM-VG based on Qwen2-VL, which is composed of three modules, including the vision encoder, projection layer, and LLM. The input images $V$ are first encoded by the vision encoder and then projected to word embedding space as $H_{v}$ through the projection layer. The visual tokens $H_{v}$ are concatenated with language embedding tokens $H_{ins}$, which are fed to the LLM to generate outputs. To handle comprehensive localization scenarios, GeM-VG uses a unified output representation for multiple object detection within multiple images.




We first conduct supervised finetuning to equip the base model with basic multi-image perception and reasoning skills. Inspired by the recent success of R1-like algorithm in both perception and reasoning tasks~\cite{deng2025openvlthinker, wang2025vl-rethinker, shen2025vlm-r1, zhan2025vision-r1, liu2025visual-rft, ma2025deepperception}, we introduce this training paradigm into the multi-image grounding task and design a rule-based reward function.
After applying reinforcement learning, the grounding performance of the model is further enhanced. The following subsections detail the data construction and reinforcement finetuning strategy.
\subsection{MG-Data-240K}

To mitigate the lack of multi-target grounding capabilities in existing multi-image grounding models, we construct a new dataset, MG-Data-240K, to support broader scenarios. This dataset addresses the limitations of existing datasets in target quantity and image relationships.
\subsubsection{Task Taxonomy}
To cover comprehensive multi-image grounding scenarios, we begin by systematically categorizing these tasks. Unlike MIG-Bench~\cite{li2025migician}, which classifies tasks based on reference requirements and forms, our taxonomy is driven by the unique cognitive demands of multi-image grounding and further refined based on image relationships. As illustrated in Figure~\ref{fig:model}(b), we divide multi-image grounding tasks into three main types based on their reliance on cross-image cues and reasoning: referential retrieval grounding, where the instructions
identify instances using explicit expression without cross-image referring; cross-image association grounding, where the targets are identified through cross-image correspondences; and reasoning grounding, which requires models to use commonsense or external knowledge to locate instances.
Within cross-image association grounding, we further divide multi-image relations into semantic, temporal, and spatial relations.
This taxonomy is informed by findings in cognitive neuroscience~\cite{baddeley2000episodic_buffer, moscovitch2006mtt}, which suggest that working memory comprises semantic, episodic, and spatial components. 
\subsubsection{Data Collection}
Previous training datasets~\cite{li2025migician, bai2025univg-r1} lack multi-target samples and mainly involve semantic relationships. To address this, we aim to expand the data with more multi-target samples, multi-view images, and diverse practical tasks. Guided by the task taxonomy, we select multiple image and video datasets as the source data.
For the raw image and video datasets we collect, we form multi-image groups by combining related images or extracting key frames from the same video sequence, according to the tasks and annotation types. Task instructions are generated using predefined templates combined with metadata such as object annotations and QA pairs from the source datasets.
Table~\ref{tab:data} summarizes the data sources and statistics. 

\begin{table}[t]
    \centering
    \fontsize{9pt}{\baselineskip}\selectfont
    \setlength{\tabcolsep}{2pt}
    \resizebox{0.95\linewidth}{!}{
    \begin{tabular}{ccl}
    \toprule
        Type & Num. & Source \\
    \midrule
        \multirow{2}{*}{Referring Retrieval Grounding} & \multirow{2}{*}{97K} & 
                                            $D^3$~\cite{xie2023d3}, \\
                                            &&COCO~\cite{lin2014mscoco} \\
    \midrule
        Semantic Association Grounding & 77K & COCO \\
    \midrule
        \multirow{2}{*}{Spatial Association Grounding} & \multirow{2}{*}{20K} & 
                                            Ego-Exo4D~\cite{grauman2024ego-exo4d}, \\
                                            &&MVTrack~\cite{xu2025mitracker} \\
    \midrule
        Temporal Association Grounding & 46K & STAR~\cite{wu2024star} \\
    \bottomrule
    \end{tabular}
    }
    \caption{Details of the constructed training dataset.}
    \label{tab:data}
\end{table}

\subsection{Reinforcement Learning for Generalized Multi-Image Grounding}
\label{sub: rl}

\begin{figure*}[ht!]
	\centering
	\begin{minipage}[t]{\linewidth}
		\centering
		\includegraphics[width=0.9\linewidth]{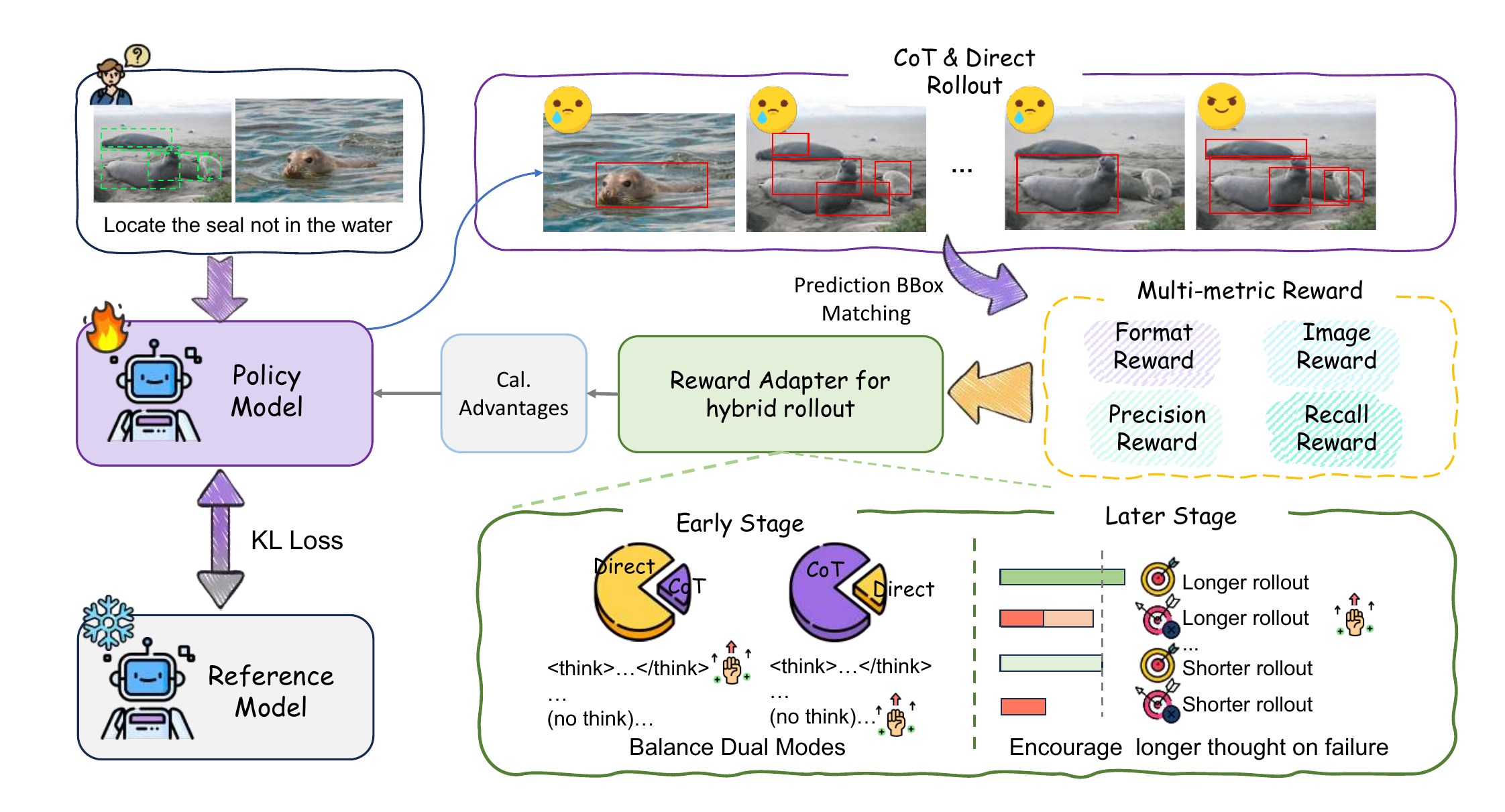}
		\caption{Framework of our reinforcement finetuning. Considering several failure types in multi-image grounding, we design a multi-dimensional rule-based reward. Before computing the advantages, we apply a reward adjustment strategy to facilitate joint optimization between both answering paradigms.} 
        \label{fig:framework}
	\end{minipage}
\end{figure*}
In this subsection, we detail the reinforcement learning algorithm used to enhance the model's generalized multi-image grounding capability. Our method is an extension of the Group Relative Policy Optimization (GRPO)~\cite{shao2024deepseekmath} algorithm to the visual grounding field.

Given an input question $q$, the policy model $\pi_\theta$ generates a group of $N$ candidate completions $\{o_1, o_2, ..., o_N\}$. For each completion $o_i$, a rule-based reward function $R(q, a, o_i)$ computes a scalar reward $r_i$, where $a$ denotes the ground truth.
To assess the relative quality of the completions in the same group, these rewards are used to compute the advantages:
\begin{equation}
    A_i = \frac{r_i - mean(\{r_j\}_{j=1}^N)}{std(\{r_j\}_{j=1}^N)}
\end{equation}
Then the policy model $\pi_\theta$ is updated by maximizing the following objective function:
\begin{equation}
    \label{eq: objective}
    \mathcal{J}_{GRPO}(\theta) = \frac{1}{N}\sum_{i=1}^N(\frac{\pi_\theta(o_i|q)}{\pi_{\theta_{old}}(o_i|q)}A_i-\beta·\mathcal{KL}(\pi_\theta(o_i|q)|\pi_{ref}(o_i|q))
\end{equation}
Where $\beta$ is a hyper-parameter to control the KL-divergence.

\subsubsection{Reward Function}
Previous methods typically formulate multi-image grounding as a single-target localization problem and adopt a simple IoU score as accuracy reward. 
To tackle more generalized scenarios, we design a reward function applicable to arbitrary numbers of instances.
As illustrated in Figure~\ref{fig:framework}, 
we propose a reward function that evaluates the output quality from multiple perspectives, incorporating format reward, image reward, precision reward and recall reward.

\begin{itemize}
	\item \textbf{Format Reward:} 
	The reward $R_{format}$ ensures that each completion $o_i$ follows a required format. Specifically, each prediction bounding box must be listed as: 
    \lstinline[breaklines=true]|Image-N:<object_ref_start>description<object_ref_start><box_start>(x1,y1),(x2,y2)<box_end>|.
    In addition, the corresponding image index must be valid numeric values within bounds. The reward is 1 if the format is satisfied and 0 otherwise.
	\item \textbf{Image Reward:}
	The reward $R_{image}$ evaluates whether the model correctly identifies which images contain the targets, regardless of the precise location of instances.
	\item \textbf{Precision Reward:}
    The precision reward assesses the quality of predicted bounding boxes at the instance level.
    Before computing rewards, we perform bipartite matching between predictions and ground-truth instances. After matching, each predicted bounding box $\hat{b}_i$ is associated with a ground-truth bounding box $b_{\text{match}(i)}$, and an Intersection over Union (IoU) score $\text{IoU}_i$.
    To encourage the model to generate more precise bounding boxes, the precision reward is defined as the average IoU score over all matched prediction instances:
    \begin{equation}
    \label{eq:precision_reward}
    R_{precision} = \frac{1}{M} \sum_{i=1}^{M} \text{IoU}_i
    \end{equation}
	\item \textbf{Recall Reward:}
	As a complement to the precision reward, the recall reward encourages the model to output all instances of interest.
    It is defined as the proportion of ground-truth instances successfully matched by a predicted box with an IoU score above a threshold $\tau$ (set to 0.5 in our experiments):
    \begin{equation}
    \label{eq:recall_reward}
    R_{recall} = \frac{1}{num(GT)} \sum_{i=1}^{M} \mathbb{I}(\text{IoU}_i \geq \tau)
    \end{equation}
\end{itemize}

The overall reward is computed as:
\begin{equation}
\label{eq:total_reward}
R = R_{format} + R_{image} + R_{precision} + R_{recall}
\end{equation}


\subsection{Reward-Modulated Hybrid Finetuning Strategy}
\label{sub: hybrid}
Multi-image grounding tasks require models to comprehend complex instructions and visual inputs. The previous approach~\cite{bai2025univg-r1} introduces explicit CoT reasoning processes and significantly improves performance on reasoning grounding tasks. However, CoT-only training can be less effective for tasks that emphasize fine-grained visual perception over complex reasoning.
For instance, as illustrated in Figure~\ref{fig:head}, the region locating task in MIG-Bench\cite{li2025migician} often involves detecting subtle visual cues or distinguishing among highly similar objects, where fine-grained perception and discrimination are critical. In such cases, ambiguous or imprecise descriptions may distract the model's attention away from critical visual information.

To leverage both paradigms, we mix CoT and direct-answer samples during SFT and remove prompts that enforce a fixed answering style.
However, we observe that the model quickly converges to direct answering during the subsequent RL training. We ascribe this to the model's reliance on CoT-specific prompts. When such prompts are removed, the model tends to direct prediction.
To mitigate this, we propose a reward-modulated hybrid finetuning strategy (Figure~\ref{fig:framework}) consisting of two components: balancing dual modes in the early training stage and encouraging longer thoughts on failure in the later stage.

\subsubsection{Balancing Dual Modes (Early Stage)} 
To avoid premature convergence to a single answering mode during training, we introduce a reward modulation mechanism based on the proportion of different modes. Completions are categorized into four types: accurate/inaccurate CoT and accurate/inaccurate direct. A completion is considered accurate if its instance-level average reward exceeds 0.5.

Let $p$ denote the proportion of CoT completions in a batch, and $\theta$ be the target balancing ratio. The adjustment magnitudes for accurate and inaccurate completions are $\delta_1$ and $\delta_2$, respectively. A scaling factor $\alpha$ controls the overall adjustment strength. The adjustment of reward is:
\begin{equation}
\label{eq:adjust_early}
    R_{\text{adjust}}(o_i) = 
    \begin{cases}
    \alpha \cdot (\theta - p) \cdot \delta_1, & \text{if } \mathbb{I}_{\text{acc-cot}}(o_i)=1 \land p < \theta \\
    \alpha \cdot (\theta - p) \cdot \delta_2, & \text{if } \mathbb{I}_{\text{inacc-cot}}(o_i)=1 \land p < \theta \\
    \alpha \cdot (p - \theta) \cdot \delta_1, & \text{if } \mathbb{I}_{\text{acc-direct}}(o_i)=1 \land p > \theta \\
    \alpha \cdot (p - \theta) \cdot \delta_2, & \text{if } \mathbb{I}_{\text{inacc-direct}}(o_i)=1 \land p > \theta \\
    0, & \text{otherwise}
    \end{cases}
\end{equation}

In our experiments, we set $\theta = 0.5$, $\alpha = 2.0$, $\delta_1 = 1.0$, and $\delta_2 = 0.5$. 
This adjustment strategy encourages a balanced behavior between both answering modes.

\subsubsection{Encouraging Longer Thoughts on Failure (Late Stage)} 
The later stage aims to improve the model’s performance across both modes. Therefore, the balancing strategy used in the early stage is removed. The optimization relies primarily on the naive reward, incentivizing the generation of responses with higher localization quality. Moreover, we find that the model finetuned using CoT data demonstrates advantages in low-scoring subtasks.
Motivated by this, an additional length-aware reward is added to the naive reward for inaccurate completions, encouraging more deliberate reasoning on lower-quality cases. The reward adjustment is computed as:
\begin{equation}
    R_{adjust}(o_i) =
    \begin{cases}
    \gamma \cdot \tilde{\ell}_i \cdot \Delta^{\max}_i, & \text{if } \mathbb{I}_{\text{inaccurate-cot}}(o_i) = 1 \\
    0, & \text{otherwise}
    \end{cases}
\end{equation}

Where $\tilde{\ell}_i = \frac{\ell_i - \min\limits_{j \in G} \ell_j}{\max\limits_{j \in G} \ell_j - \min\limits_{j \in G} \ell_j + \epsilon}$ is the normalized length of the completion $o_i$ within the group $G$. To ensure that the adjusted rewards of inaccurate completions do not exceed that of accurate ones, the adjustment is capped by $\Delta^{\max}_i = \max(0, r^{\text{acc}}_{\min} - r_i)$, where $r^{\text{acc}}_{\min}$ is the lowest naive reward of accurate completions within the group. The scaling factor $\gamma$ is set to $0.3$ for soft modulation.

During RL optimization, we use the adjusted rewards $R_{adjusted} = R + R_{adjust}$ to calculate advantages.





\section{Experiments}
\begin{table*}[ht]
\centering
\small
\resizebox{0.95\linewidth}{!}{
\begin{tabular}{l|ccc|cccc|ccc|c}
\toprule
\multirow{3}{*}[-1.4ex]{Model} & \multicolumn{3}{c|}{Spontaneous Grounding} & \multicolumn{7}{c|}{Referential Grounding}&\multirow{3}{*}[-1.4ex]{AVG} \\
\cmidrule(lr){2-4} \cmidrule(lr){5-11}
& \multicolumn{2}{c|}{Difference}& Similarity & \multicolumn{4}{c|}{Visual Reference} &  \multicolumn{1}{c|}{Textual} & \multicolumn{2}{c|}{Visual+Textual} \\
\cmidrule(lr){2-3} \cmidrule(lr){4-4} \cmidrule(lr){5-8} \cmidrule(lr){9-9} \cmidrule(lr){10-11}
& Static & \multicolumn{1}{c|}{Robust} & Common & OT & MV & Region & Refer & GG & Reason & Co-Re \\
\midrule
Qwen2-VL-72B~\cite{wang2024qwen2vl} & 51.13 & 43.61 & 73.74 & 24.54 & 32.63 & 19.86 & 37.37 & 67.83 & 50.51 & 17.94 & 41.91 \\
\midrule
Mantis~\cite{jiang2024mantis} & 1.52 & 0.00 & 3.31 & 12.18 & 2.08 & 1.00 & 1.01 & 10.02 & 0.00 & 0.85 & 3.20 \\
LLaVA-OV-7B~\cite{li2024llava-onevision} & 6.06 & 3.19 & 3.43 & 0.18 & 1.04 & 1.08 & 9.09 & 15.43 & 6.93 & 0.85 & 4.73 \\
Minicpm2.6~\cite{yao2024minicpm-v} & 14.58 & 2.13 & 14.34 & 9.82 & 2.65 & 1.75 & 11.11 & 20.62 & 2.97 & 2.56 & 7.55 \\
mPLUG-Owl3~\cite{ye2024mplug-owl3} & 18.56 & 6.38 & 34.93 & 8.55 & 7.64 & 2.41 & 7.07 & 22.85 & 9.09 & 5.98 & 12.35 \\
InternVL2-8B~\cite{team2024internvl2} & 8.52 & 19.15 & 38.40 & 19.82 & 10.07 & 5.24 & 34.34 & 39.79 & 26.80 & 7.69 & 20.98 \\
Qwen2-VL-7B~\cite{wang2024qwen2vl} & 29.92 & 36.17 & 43.07 & 14.55 & 9.38 & 15.54 & 29.29 & 63.51 & 44.33 & 14.30 & 33.03 \\
Migician~\cite{li2025migician} & 70.64 & 45.74 & 72.76 & 67.82 & 60.07 & 72.57 & 75.76 & 84.12 & 52.58 & 33.33 & 63.54 \\
UniVG-R1~\cite{bai2025univg-r1} & 71.97 & \textbf{58.51} & \textbf{93.13} & 76.36 & 66.32 & 81.71 & 82.83 & \textbf{88.04} & 62.89 & \textbf{44.44} & 72.64 \\
\midrule
\textbf{GeM-VG} & \textbf{76.89} & 56.38 & 91.53 & \textbf{86.55} & \textbf{68.75} & \textbf{85.70} & \textbf{84.85} & 86.80 & \textbf{68.04} & 41.03 & \textbf{74.65} \\
\bottomrule
\end{tabular}}
\caption{Performance on MIG-Bench. OT, MV, GG and Co-Re respectively means object tracking, multi-view grounding, group grounding and correspondence. The metric is Acc@0.5, which considers a prediction correct if its IoU with the ground truth exceeds 0.5.}
\label{tab:mig-bench}
\end{table*}

\begin{table}[t!]
\centering
\resizebox{0.95\linewidth}{!}{
\begin{tabular}{l|ccc|c}
\toprule
Model & $AP_{50}^{ref}$ & $AP_{50}^{com}$ & $AP_{50}^{rea}$ & $AP_{50}$ \\
\midrule
Gemini-1.5 Pro~\cite{team2024gemini1.5} & 30.5 & 30.0 & 26.1 & 28.4 \\
CogVLM-G~\cite{wang2024cogvlm} & 21.1 & 19.0 & 16.5 & 18.2 \\
Qwen2-VL-7B~\cite{wang2024qwen2vl} & 22.7 & 22.4 & 17.2 & 20.2 \\
Migician~\cite{li2025migician} & 20.3 & 26.4 & 20.1 & 23.1 \\
\textbf{GeM-VG} & \textbf{42.0} & \textbf{33.6} & \textbf{28.7} & \textbf{32.8} \\
\bottomrule
\end{tabular}}
\caption{Performance comparison on MC-Bench. The superscripts ref, com and rea denote the results for referring, comparison, reasoning instruction type respectively.} 
\label{tab:mc-bench}
\end{table}

\begin{table}[t!]
\centering
\resizebox{1.0\linewidth}{!}{
\begin{tabular}{l|cc|cc}
\toprule
\multirow{2}{*}{Model} & \multicolumn{2}{|c|}{Single image Grounding} & \multicolumn{2}{|c}{Video Grounding}\\
\cmidrule(lr){2-3} \cmidrule(lr){4-5}
 & ODINW & LLMSeg & ReasonVOS & ReVOS \\
\midrule
Qwen2-VL-7B~\cite{wang2024qwen2vl} & 32.0 & 35.53 & 9.83 & 23.55 \\
Qwen2.5-VL-7B~\cite{bai2025qwen25vl} & 37.0 & — & — & — \\
Migician~\cite{li2025migician} & 21.9 & 34.68 & 33.41 & 39.70 \\
UniVG-R1~\cite{bai2025univg-r1} & 33.3 & 50.60 & 58.73 & 60.03 \\
\textbf{GeM-VG} & \textbf{41.1} & \textbf{50.90} & \textbf{64.41} & \textbf{60.52} \\
\bottomrule
\end{tabular}}
\caption{Single image and video grounding results. For ODINW, we follow the evaluation setting in~\cite{bai2025qwen25vl}. For video grounding, we follow the setting in~\cite{bai2025univg-r1}}
\label{tab:single image and video grounding}
\end{table}

\begin{table}[t!]
\centering
\resizebox{1.0\linewidth}{!}{
\begin{tabular}{c|cccc}
\toprule
Model & MuirBench & BLINK val & MIBench & MMIU \\ 
\midrule
LLaVA-1.5~\cite{liu2024llava1.5} & 23.46 & 37.13 & 26.83 & 19.20 \\
CogVLM~\cite{wang2024cogvlm} & 20.85 & 41.54 & — & 23.57 \\
Idefics2-8B~\cite{laurenccon2024idefics2} & 26.08 & — & 46.39 & 27.80 \\
mPLUG-Owl3~\cite{ye2024mplug-owl3} & 39.67 & 50.30 & 56.66 & 21.72 \\  
InternVL2-8B~\cite{team2024internvl2} & 48.70 & 50.57 & 52.91 & 42.00  \\ 
Mantis~\cite{jiang2024mantis} & 44.50 & 49.05 & 45.09 & 45.60 \\
LLaVA-OV-7B~\cite{li2024llava-onevision} & 41.80 & 48.20 & 71.29 &  44.46 \\ 
MiniCPM-V 2.6~\cite{yao2024minicpm-v} 2.6 & 42.65 & 51.45 & 71.09 & 50.19 \\ 
Qwen2-VL-7B~\cite{wang2024qwen2vl} & 39.88 & 52.35 & 68.06 & 54.36 \\
Migician~\cite{li2025migician} & 57.81 & 51.53 & \textbf{71.42} & 54.89 \\ 
UniVG-R1~\cite{bai2025univg-r1} & 44.77 & 51.55 & 67.29 & 53.03 \\ 
\midrule 
\textbf{GeM-VG} & \textbf{58.20} & \textbf{52.97} & 70.16 & \textbf{55.01} \\ 
\bottomrule
\end{tabular}}
\caption{Multi-image understanding results.}
\label{tab:multi-image understanding}
\end{table}

\begin{table}[t!]
\centering
\resizebox{0.95\linewidth}{!}{
\begin{tabular}{cc|cccc}
\toprule
 \multirow{2}{*}{precision} & \multirow{2}{*}{recall} & \multicolumn{4}{c}{ODINW} \\
 \cmidrule(lr){3-6}
 &&AerialDrone & Aquarium & EgoHands & thermal\\
\midrule
 \multicolumn{2}{c|}{Baseline} & 2.3 & 23.7 & 48.7 & 40.5\\
  & $\checkmark$ & 3.3 & 19.1 & 49.3 & 41.1\\
 $\checkmark$ &  & 2.1 & 17.3 & 25.2 & \textbf{49.3}\\
 $\checkmark$ & $\checkmark$ & \textbf{5.3} & \textbf{25.6} & \textbf{56.6} & 45.4\\
\bottomrule
\end{tabular}
}
\caption{Ablation on Reward Design.}
\label{tab:ablation reward}
\end{table}

\begin{table*}[ht]
\centering
\small
\resizebox{0.95\linewidth}{!}{
\begin{tabular}{l|ccc|cccc|ccc|c}
\toprule
\multirow{3}{*}[-1.4ex]{Models} & \multicolumn{3}{c|}{Spontaneous Grounding} & \multicolumn{7}{c|}{Referential Grounding}&\multirow{3}{*}{AVG} \\
\cmidrule(lr){2-4} \cmidrule(lr){5-11}
& \multicolumn{2}{c|}{Difference}& Similarity & \multicolumn{4}{c|}{Visual Reference} &  \multicolumn{1}{c|}{Textual} & \multicolumn{2}{c|}{Visual+Textual} \\
\cmidrule(lr){2-3} \cmidrule(lr){4-4} \cmidrule(lr){5-8} \cmidrule(lr){9-9} \cmidrule(lr){10-11}
& Static & \multicolumn{1}{c|}{Robust} & Common & OT & MV & Region & Refer & GG & Reason & Co-Re \\
\midrule
Qwen2-VL-7B~\cite{wang2024qwen2vl} & 29.92 & 36.17 & 43.07 & 14.55 & 9.38 & 15.54 & 29.29 & 63.51 & 44.33 & 16.24 & 30.20 \\
\cmidrule(lr){1-12}
\multicolumn{12}{c}{\textbf{SFT (Stage 2)}} \\

\cmidrule(lr){1-12}
CoT & 76.52 & 51.06 & 89.33 & 80.18 & 68.06 & 80.63 & 80.81 & 85.77 & \textbf{62.89} & \textbf{39.32} & 71.46 \\
Direct & 76.70 & 50.00 & 89.20 & 82.91 & 69.10 & 86.03 & 80.81 & 87.84 & 45.36 & 26.50 & 69.45 \\
Mix & 73.48 & 52.13 & 90.67 & 83.64 & 68.06 & 82.46 & 82.83 & 87.84 & 61.86 & 36.75 & 71.97 \\
\cmidrule(lr){1-12}

\multicolumn{12}{c}{\textbf{RL (Stage 3)}} \\
\cmidrule(lr){1-12}
w/o mode balancing & 77.08 & 55.32 & 89.08 & 86.73 & 70.41 & \textbf{86.87} & 82.83 & 87.84 & 51.55 & 31.62 & 71.91 \\
w/o late stage strategy & \textbf{79.92} & 51.06 & \textbf{92.39} & \textbf{86.91} & 69.44 & 85.95 & \textbf{85.86} & 87.63 & 52.58 & 32.48 & 72.4 \\
only CoT & 77.08 & 54.26 & 90.92 & 86.55 & \textbf{71.53} & 85.95 & 84.85 & 87.22 & 59.79 & 37.61 & 73.58 \\
early stage+late stage & 76.89 & \textbf{56.38} & 91.53 & 85.82 & \textbf{71.53} & 86.20 & \textbf{85.86} & \textbf{88.66} & \textbf{62.89} & 37.61 & \textbf{74.31} \\
\bottomrule
\end{tabular}}
\caption{Ablation on Hybrid Finetuning Strategy.}
\label{tab:ablation hybrid}
\end{table*}

\subsection{Implementation Details}
\subsubsection{Training Data}
The training process comprises three stages. To equip the model with comprehensive multi-image grounding capabilities, we combine the MGrounding-630k dataset with an additional 240k samples constructed from multiple existing datasets to form the training data for Stage1-SFT. To mitigate catastrophic forgetting, we also mix multi-image understanding and single-image grounding data into the training. In Stage2-SFT, we incorporate both CoT and direct-answer annotations derived from UniVG-R1~\cite{bai2025univg-r1}. The RL stage is trained on 26k samples, which are sampled from prior stages.

\subsubsection{Benchmarks}
We conduct comprehensive evaluations across multi-image grounding, single-image grounding, video grounding, and multi-image understanding tasks. For multi-image grounding, we evaluate our model on MIG-Bench and MC-Bench. Among them, MIG-Bench requires localizing a single region in the target image, while MC-Bench involves grounding an unspecified number of instances. For single-image grounding, we evaluate on ODINW~\cite{li2022odinw}, which contains rare categories in real-world scenarios, and LLMSeg~\cite{wang2024llmseg}, a reasoning grounding benchmark. For video grounding, we uniformly sample 6 frames and task the model with localizing within one of these frames, with evaluations conducted on ReasonVOS~\cite{bai2024reasonvos} and ReVOS~\cite{yan2024revos}. To assess the model’s general multi-image understanding capabilities, we further incorporate several representative benchmarks including MMIU~\cite{meng2024mmiu}, MuirBench~\cite{wang2024muirbench}, MIBench\cite{liu2024mibench} and BLINK~\cite{fu2024blink}.

\subsubsection{Training Configurations}
We adopt Qwen2-VL-7B \cite{wang2024qwen2vl} as the base model due to its strong multi-image understanding and single-image grounding capabilities. During the SFT stage, we employ a learning rate of 2e-6 and a total batch size of 128. For RL training, we use the multimodal version of the Open-R1 framework~\cite{chen2025r1-v} with its default configuration. The learning rate is set to 1e-6, and the batch size is 16. For each prompt, 8 completions are sampled, with a maximum completion length of 1024. The training is conducted over 1 epoch on 16 NVIDIA H100(80G) GPUs, consuming approximately 8, 2 and 10 hours for each stage, respectively.

\subsection{Main Results}
\subsubsection{Multi-image Grounding}
To comprehensively evaluate the model’s multi-image grounding capability, we conduct experiments on MIG-Bench~\cite{li2025migician} and MC-Bench~\cite{xu2024mc-bench}. MIG-Bench is a free-form multi-image grounding benchmark that encompasses diverse subtasks, requiring the model to localize a single region of interest in the target image based on visual or textual references. As shown in Table~\ref{tab:mig-bench}, our method achieves the best overall performance. Compared to the previous leading model, GeM-VG shows improvements on subtasks that require discerning perception, such as static difference and region locating, as well as those involving reasoning, such as reasoning grounding.
Unlike MIG-Bench, MC-Bench evaluates the model’s ability to identify an arbitrary number of instances that match the instruction. The textual instructions in MC-Bench are categorized into three styles: referring, comparison, and reasoning. As shown in Table~\ref{tab:mc-bench}, our model outperforms the base model and Migician by 12.6\% and 9.7\% respectively, demonstrating superior capability in grounding multiple instances in multi-image scenarios.

\subsubsection{Single-image/Video Grounding}
We also evaluate our model on single-image and video grounding tasks. As shown in Table~\ref{tab:single image and video grounding}, our model achieves strong performance across all four benchmarks. Specifically, it outperforms UniVG-R1 by 7.8\% on ODINW and 5.68\% on ReasonVOS, demonstrating superior multi-object localization, temporal understanding, and reasoning capabilities. More details about the results are provided in the supplements.

\subsubsection{Multi-image Understanding}
Our model not only excels in multi-image grounding, but also achieves competitive performance on general multi-image understanding tasks, as shown in Table~\ref{tab:multi-image understanding}. While UniVG-R1 significantly improves over previous methods on multi-image grounding, it falls short of Migician in broader multi-image understanding. In contrast, our model consistently achieves superior results across a range of benchmarks. Notably, it outperforms other models on the counting and visual grounding subtasks of MuirBench, demonstrating the effectiveness of our approach for generalized multi-image grounding. More detailed results are demonstrated in the supplements.

\subsection{Ablation Studies}

\subsubsection{Effectiveness of Reward Design}
We conduct ablation experiments to investigate the impact of different components in our reward design. Among them, the format reward guides the model to generate outputs in the expected format for consistent localization results parsing, while the image reward provides coarse image-level feedback without considering the quality of predicted bounding boxes. Therefore, our primary focus is on the effects of the precision and recall rewards. Experiments are conducted on several representative ODINW subsets, encompassing a range of challenging scenarios, such as dense objects, small-scale targets, occlusions, and rare image domains.
As shown in Table~\ref{tab:ablation reward}, when relying solely on the precision reward, performance degrades significantly on subsets like Aquarium (dense targets) and EgoHands (occlusions and shape variations), as these scenarios are prone to missed detections and the model lacks incentive to locate all mentioned instances. Conversely, when excluding the precision reward, the absence of fine-grained supervision on box quality leads to an increase in low-quality predictions, resulting in decreased AP metrics in some cases.
By incorporating all reward components, the model achieves consistent improvements across all test sets, demonstrating the effectiveness of our reward design.

\subsubsection{Effectiveness of Hybrid Finetuning Strategy}
The hybrid finetuning strategy serves as a mechanism to enhance the overall performance of the model across various grounding tasks. For the SFT stage, we compare the performance of models trained with different modes of data on MIG-Bench. As shown in Table~\ref{tab:ablation hybrid}, using pure CoT data significantly improves performance on reasoning-intensive tasks such as Reason and Co-Re, compared to direct-answer data. However, this comes at the cost of reduced performance on perception-oriented tasks like Region. In contrast, combining both paradigms for finetuning yields better overall performance, which is adopted as our training recipe.

In the RL stage, we conduct an ablation study to compare the impact of different reward modulation strategies. We initially conduct RL training without any reward intervention, and find that the completions tend to collapse early into direct-answer mode. As depicted in Table~\ref{tab:ablation hybrid}, this results in degraded performance on reasoning-intensive subtasks and limited overall improvement. After adopting a dual-mode balancing strategy throughout training, the model achieves improved performance compared to no intervention, demonstrating the benefit of introducing reasoning chains during exploration. Ultimately, by applying a two-stage reward adjustment—balancing both completion modes in early training and shifting toward more thoughts on failure in later stage, the model achieves the best overall performance, outperforming all other variants including the pure CoT mode. These results validate the effectiveness of our hybrid finetuning approach in achieving robust performance across diverse multi-image grounding tasks. 

\section{Conclusion}
In this paper, we propose GeM-VG, an MLLM capable of generalized multi-image grounding while retaining strong capabilities on single-image grounding and multi-image understanding. To endow and incentivize the base model with multi-image grounding and reasoning capabilities, we introduce the MG-Data-240K dataset and a hybrid finetuning strategy based on R1-like reinforcement learning. 
Extensive experiments demonstrate that our model achieves superior performance across multi-image grounding, single-image grounding and multi-image understanding benchmarks. We hope this work will encourage further research into developing MLLMs with advanced generalized grounding capabilities to support a wider range of real-world applications.


\setcounter{page}{1}
{
    \bibliography{aaai2026}

\begin{thebibliography}{57}
\providecommand{\natexlab}[1]{#1}

\bibitem[{Bai et~al.(2025{\natexlab{a}})Bai, Chen, Liu, Wang, Ge, Song, Dang, Wang, Wang, Tang et~al.}]{bai2025qwen25vl}
Bai, S.; Chen, K.; Liu, X.; Wang, J.; Ge, W.; Song, S.; Dang, K.; Wang, P.; Wang, S.; Tang, J.; et~al. 2025{\natexlab{a}}.
\newblock Qwen2. 5-vl technical report.
\newblock \emph{arXiv preprint arXiv:2502.13923}.

\bibitem[{Bai et~al.(2025{\natexlab{b}})Bai, Li, Liu, Tang, Zhang, Sun, Chu, and Tang}]{bai2025univg-r1}
Bai, S.; Li, M.; Liu, Y.; Tang, J.; Zhang, H.; Sun, L.; Chu, X.; and Tang, Y. 2025{\natexlab{b}}.
\newblock Univg-r1: Reasoning guided universal visual grounding with reinforcement learning.
\newblock \emph{arXiv preprint arXiv:2505.14231}.

\bibitem[{Bai et~al.(2024)Bai, He, Mei, Wang, Gao, Chen, Zhang, and Shou}]{bai2024reasonvos}
Bai, Z.; He, T.; Mei, H.; Wang, P.; Gao, Z.; Chen, J.; Zhang, Z.; and Shou, M.~Z. 2024.
\newblock One token to seg them all: Language instructed reasoning segmentation in videos.
\newblock \emph{Advances in Neural Information Processing Systems}, 37: 6833--6859.

\bibitem[{Chen et~al.(2023)Chen, Zhang, Zeng, Zhang, Zhu, and Zhao}]{chen2023shikra}
Chen, K.; Zhang, Z.; Zeng, W.; Zhang, R.; Zhu, F.; and Zhao, R. 2023.
\newblock Shikra: Unleashing multimodal llm's referential dialogue magic.
\newblock \emph{arXiv preprint arXiv:2306.15195}.

\bibitem[{Chen et~al.(2025)Chen, Li, Zhao, and Song}]{chen2025r1-v}
Chen, L.; Li, L.; Zhao, H.; and Song, Y. 2025.
\newblock Vinci. R1-v: Reinforcing super generalization ability in vision-language models with less than \$3.

\bibitem[{Chen et~al.(2024)Chen, Wang, Cao, Liu, Gao, Cui, Zhu, Ye, Tian, Liu et~al.}]{chen2024internvl2.5}
Chen, Z.; Wang, W.; Cao, Y.; Liu, Y.; Gao, Z.; Cui, E.; Zhu, J.; Ye, S.; Tian, H.; Liu, Z.; et~al. 2024.
\newblock Expanding performance boundaries of open-source multimodal models with model, data, and test-time scaling.
\newblock \emph{arXiv preprint arXiv:2412.05271}.

\bibitem[{Deng et~al.(2025)Deng, Bansal, Yin, Peng, Wang, and Chang}]{deng2025openvlthinker}
Deng, Y.; Bansal, H.; Yin, F.; Peng, N.; Wang, W.; and Chang, K.-W. 2025.
\newblock Openvlthinker: An early exploration to complex vision-language reasoning via iterative self-improvement.
\newblock \emph{arXiv preprint arXiv:2503.17352}.

\bibitem[{Fu et~al.(2024)Fu, Hu, Li, Feng, Wang, Lin, Roth, Smith, Ma, and Krishna}]{fu2024blink}
Fu, X.; Hu, Y.; Li, B.; Feng, Y.; Wang, H.; Lin, X.; Roth, D.; Smith, N.~A.; Ma, W.-C.; and Krishna, R. 2024.
\newblock Blink: Multimodal large language models can see but not perceive.
\newblock In \emph{European Conference on Computer Vision}, 148--166. Springer.

\bibitem[{Grauman et~al.(2024)Grauman, Westbury, Torresani, Kitani, Malik, Afouras, Ashutosh, Baiyya, Bansal, Boote et~al.}]{grauman2024ego-exo4d}
Grauman, K.; Westbury, A.; Torresani, L.; Kitani, K.; Malik, J.; Afouras, T.; Ashutosh, K.; Baiyya, V.; Bansal, S.; Boote, B.; et~al. 2024.
\newblock Ego-exo4d: Understanding skilled human activity from first-and third-person perspectives.
\newblock In \emph{Proceedings of the IEEE/CVF Conference on Computer Vision and Pattern Recognition}, 19383--19400.

\bibitem[{Guo et~al.(2025)Guo, Yang, Zhang, Song, Zhang, Xu, Zhu, Ma, Wang, Bi et~al.}]{guo2025deepseek-r1}
Guo, D.; Yang, D.; Zhang, H.; Song, J.; Zhang, R.; Xu, R.; Zhu, Q.; Ma, S.; Wang, P.; Bi, X.; et~al. 2025.
\newblock Deepseek-r1: Incentivizing reasoning capability in llms via reinforcement learning.
\newblock \emph{arXiv preprint arXiv:2501.12948}.

\bibitem[{Huang et~al.(2025)Huang, Jia, Zhai, Cao, Ye, Zhao, Xu, Hu, and Lin}]{huang2025vision-r1}
Huang, W.; Jia, B.; Zhai, Z.; Cao, S.; Ye, Z.; Zhao, F.; Xu, Z.; Hu, Y.; and Lin, S. 2025.
\newblock Vision-r1: Incentivizing reasoning capability in multimodal large language models.
\newblock \emph{arXiv preprint arXiv:2503.06749}.

\bibitem[{Jaech et~al.(2024)Jaech, Kalai, Lerer, Richardson, El-Kishky, Low, Helyar, Madry, Beutel, Carney et~al.}]{jaech2024openai-o1}
Jaech, A.; Kalai, A.; Lerer, A.; Richardson, A.; El-Kishky, A.; Low, A.; Helyar, A.; Madry, A.; Beutel, A.; Carney, A.; et~al. 2024.
\newblock Openai o1 system card.
\newblock \emph{arXiv preprint arXiv:2412.16720}.

\bibitem[{Jiang et~al.(2024)Jiang, He, Zeng, Wei, Ku, Liu, and Chen}]{jiang2024mantis}
Jiang, D.; He, X.; Zeng, H.; Wei, C.; Ku, M.; Liu, Q.; and Chen, W. 2024.
\newblock Mantis: Interleaved multi-image instruction tuning.
\newblock \emph{arXiv preprint arXiv:2405.01483}.

\bibitem[{Kamath et~al.(2021)Kamath, Singh, LeCun, Synnaeve, Misra, and Carion}]{kamath2021mdetr}
Kamath, A.; Singh, M.; LeCun, Y.; Synnaeve, G.; Misra, I.; and Carion, N. 2021.
\newblock Mdetr-modulated detection for end-to-end multi-modal understanding.
\newblock In \emph{Proceedings of the IEEE/CVF international conference on computer vision}, 1780--1790.

\bibitem[{Lauren{\c{c}}on et~al.(2024)Lauren{\c{c}}on, Tronchon, Cord, and Sanh}]{laurenccon2024idefics2}
Lauren{\c{c}}on, H.; Tronchon, L.; Cord, M.; and Sanh, V. 2024.
\newblock What matters when building vision-language models?
\newblock \emph{Advances in Neural Information Processing Systems}, 37: 87874--87907.

\bibitem[{Li et~al.(2024{\natexlab{a}})Li, Zhang, Guo, Zhang, Li, Zhang, Zhang, Zhang, Li, Liu et~al.}]{li2024llava-onevision}
Li, B.; Zhang, Y.; Guo, D.; Zhang, R.; Li, F.; Zhang, H.; Zhang, K.; Zhang, P.; Li, Y.; Liu, Z.; et~al. 2024{\natexlab{a}}.
\newblock Llava-onevision: Easy visual task transfer.
\newblock \emph{arXiv preprint arXiv:2408.03326}.

\bibitem[{Li et~al.(2024{\natexlab{b}})Li, Zhang, Zhang, Zhang, Li, Li, Ma, and Li}]{li2024llava-next-interleave}
Li, F.; Zhang, R.; Zhang, H.; Zhang, Y.; Li, B.; Li, W.; Ma, Z.; and Li, C. 2024{\natexlab{b}}.
\newblock Llava-next-interleave: Tackling multi-image, video, and 3d in large multimodal models.
\newblock \emph{arXiv preprint arXiv:2407.07895}.

\bibitem[{Li et~al.(2022)Li, Zhang, Zhang, Yang, Li, Zhong, Wang, Yuan, Zhang, Hwang et~al.}]{li2022odinw}
Li, L.~H.; Zhang, P.; Zhang, H.; Yang, J.; Li, C.; Zhong, Y.; Wang, L.; Yuan, L.; Zhang, L.; Hwang, J.-N.; et~al. 2022.
\newblock Grounded language-image pre-training.
\newblock In \emph{Proceedings of the IEEE/CVF conference on computer vision and pattern recognition}, 10965--10975.

\bibitem[{Li et~al.(2025)Li, Huang, Chen, Huang, Huang, Guo, Liu, Xu, Li, Li et~al.}]{li2025migician}
Li, Y.; Huang, H.; Chen, C.; Huang, K.; Huang, C.; Guo, Z.; Liu, Z.; Xu, J.; Li, Y.; Li, R.; et~al. 2025.
\newblock Migician: Revealing the magic of free-form multi-image grounding in multimodal large language models.
\newblock \emph{arXiv preprint arXiv:2501.05767}.

\bibitem[{Lin et~al.(2014)Lin, Maire, Belongie, Hays, Perona, Ramanan, Doll{\'a}r, and Zitnick}]{lin2014mscoco}
Lin, T.-Y.; Maire, M.; Belongie, S.; Hays, J.; Perona, P.; Ramanan, D.; Doll{\'a}r, P.; and Zitnick, C.~L. 2014.
\newblock Microsoft coco: Common objects in context.
\newblock In \emph{European conference on computer vision}, 740--755. Springer.

\bibitem[{Liu, Ding, and Jiang(2023)}]{grefcoco}
Liu, C.; Ding, H.; and Jiang, X. 2023.
\newblock {GRES}: Generalized Referring Expression Segmentation.
\newblock In \emph{CVPR}.

\bibitem[{Liu et~al.(2024{\natexlab{a}})Liu, Li, Li, and Lee}]{liu2024llava1.5}
Liu, H.; Li, C.; Li, Y.; and Lee, Y.~J. 2024{\natexlab{a}}.
\newblock Improved baselines with visual instruction tuning.
\newblock In \emph{Proceedings of the IEEE/CVF conference on computer vision and pattern recognition}, 26296--26306.

\bibitem[{Liu et~al.(2024{\natexlab{b}})Liu, Zhang, Xu, Shi, Jiang, Yan, Zhang, Huang, Yuan, Li et~al.}]{liu2024mibench}
Liu, H.; Zhang, X.; Xu, H.; Shi, Y.; Jiang, C.; Yan, M.; Zhang, J.; Huang, F.; Yuan, C.; Li, B.; et~al. 2024{\natexlab{b}}.
\newblock Mibench: Evaluating multimodal large language models over multiple images.
\newblock \emph{arXiv preprint arXiv:2407.15272}.

\bibitem[{Liu et~al.(2024{\natexlab{c}})Liu, Zeng, Ren, Li, Zhang, Yang, Jiang, Li, Yang, Su et~al.}]{liu2024groundingdino}
Liu, S.; Zeng, Z.; Ren, T.; Li, F.; Zhang, H.; Yang, J.; Jiang, Q.; Li, C.; Yang, J.; Su, H.; et~al. 2024{\natexlab{c}}.
\newblock Grounding dino: Marrying dino with grounded pre-training for open-set object detection.
\newblock In \emph{European conference on computer vision}, 38--55. Springer.

\bibitem[{Liu et~al.(2025)Liu, Sun, Zang, Dong, Cao, Duan, Lin, and Wang}]{liu2025visual-rft}
Liu, Z.; Sun, Z.; Zang, Y.; Dong, X.; Cao, Y.; Duan, H.; Lin, D.; and Wang, J. 2025.
\newblock Visual-rft: Visual reinforcement fine-tuning.
\newblock \emph{arXiv preprint arXiv:2503.01785}.

\bibitem[{Ma et~al.(2025)Ma, Ding, Luo, Chen, Guo, Wong, Feng, and Sun}]{ma2025deepperception}
Ma, X.; Ding, Z.; Luo, Z.; Chen, C.; Guo, Z.; Wong, D.~F.; Feng, X.; and Sun, M. 2025.
\newblock Deepperception: Advancing r1-like cognitive visual perception in mllms for knowledge-intensive visual grounding.
\newblock \emph{arXiv preprint arXiv:2503.12797}.

\bibitem[{Meng et~al.(2024)Meng, Wang, Li, Lu, Tian, Liao, Zhu, Dai, Qiao, Luo et~al.}]{meng2024mmiu}
Meng, F.; Wang, J.; Li, C.; Lu, Q.; Tian, H.; Liao, J.; Zhu, X.; Dai, J.; Qiao, Y.; Luo, P.; et~al. 2024.
\newblock Mmiu: Multimodal multi-image understanding for evaluating large vision-language models.
\newblock \emph{arXiv preprint arXiv:2408.02718}.

\bibitem[{Moscovitch et~al.(2006)Moscovitch, Nadel, Winocur, Gilboa, and Rosenbaum}]{moscovitch2006mtt}
Moscovitch, M.; Nadel, L.; Winocur, G.; Gilboa, A.; and Rosenbaum, R.~S. 2006.
\newblock The cognitive neuroscience of remote episodic, semantic and spatial memory.
\newblock \emph{Current opinion in neurobiology}, 16(2): 179--190.

\bibitem[{Nagaraja, Morariu, and Davis(2016)}]{nagaraja2016refcocog}
Nagaraja, V.~K.; Morariu, V.~I.; and Davis, L.~S. 2016.
\newblock Modeling context between objects for referring expression understanding.
\newblock In \emph{European Conference on Computer Vision}, 792--807. Springer.

\bibitem[{OpenAI(2023)}]{openai2023gpt4o}
OpenAI, R. 2023.
\newblock Gpt-4 technical report. arxiv 2303.08774.
\newblock \emph{View in Article}, 2(5): 1.

\bibitem[{Peng et~al.(2023)Peng, Wang, Dong, Hao, Huang, Ma, and Wei}]{peng2023kosmos-2}
Peng, Z.; Wang, W.; Dong, L.; Hao, Y.; Huang, S.; Ma, S.; and Wei, F. 2023.
\newblock Kosmos-2: Grounding multimodal large language models to the world.
\newblock \emph{arXiv preprint arXiv:2306.14824}.

\bibitem[{Plummer et~al.(2015)Plummer, Wang, Cervantes, Caicedo, Hockenmaier, and Lazebnik}]{plummer2015flickr30k}
Plummer, B.~A.; Wang, L.; Cervantes, C.~M.; Caicedo, J.~C.; Hockenmaier, J.; and Lazebnik, S. 2015.
\newblock Flickr30k entities: Collecting region-to-phrase correspondences for richer image-to-sentence models.
\newblock In \emph{Proceedings of the IEEE international conference on computer vision}, 2641--2649.

\bibitem[{Shao et~al.(2024)Shao, Wang, Zhu, Xu, Song, Bi, Zhang, Zhang, Li, Wu et~al.}]{shao2024deepseekmath}
Shao, Z.; Wang, P.; Zhu, Q.; Xu, R.; Song, J.; Bi, X.; Zhang, H.; Zhang, M.; Li, Y.; Wu, Y.; et~al. 2024.
\newblock Deepseekmath: Pushing the limits of mathematical reasoning in open language models.
\newblock \emph{arXiv preprint arXiv:2402.03300}.

\bibitem[{Shen et~al.(2025)Shen, Liu, Li, Fang, Ma, Liao, Shen, Zhang, Zhao, Zhang et~al.}]{shen2025vlm-r1}
Shen, H.; Liu, P.; Li, J.; Fang, C.; Ma, Y.; Liao, J.; Shen, Q.; Zhang, Z.; Zhao, K.; Zhang, Q.; et~al. 2025.
\newblock Vlm-r1: A stable and generalizable r1-style large vision-language model.
\newblock \emph{arXiv preprint arXiv:2504.07615}.

\bibitem[{Team et~al.(2024{\natexlab{a}})Team, Anil, Borgeaud, Wu, Alayrac, Yu, Soricut, Schalkwyk, Dai, Hauth et~al.}]{team2024geminiPRO}
Team, G.; Anil, R.; Borgeaud, S.; Wu, Y.; Alayrac, J.; Yu, J.; Soricut, R.; Schalkwyk, J.; Dai, A.; Hauth, A.; et~al. 2024{\natexlab{a}}.
\newblock Gemini: A family of highly capable multimodal models, 2024.
\newblock \emph{arXiv preprint arXiv:2312.11805}, 10.

\bibitem[{Team et~al.(2024{\natexlab{b}})Team, Georgiev, Lei, Burnell, Bai, Gulati, Tanzer, Vincent, Pan, Wang et~al.}]{team2024gemini1.5}
Team, G.; Georgiev, P.; Lei, V.~I.; Burnell, R.; Bai, L.; Gulati, A.; Tanzer, G.; Vincent, D.; Pan, Z.; Wang, S.; et~al. 2024{\natexlab{b}}.
\newblock Gemini 1.5: Unlocking multimodal understanding across millions of tokens of context.
\newblock \emph{arXiv preprint arXiv:2403.05530}.

\bibitem[{Team(2024)}]{team2024internvl2}
Team, O. 2024.
\newblock Internvl2: Better than the best—expanding performance boundaries of open-source multimodal models with the progressive scaling strategy.

\bibitem[{Wang et~al.(2024{\natexlab{a}})Wang, Fu, Huang, Li, Liu, Liu, Ma, Xu, Zhou, Zhang et~al.}]{wang2024muirbench}
Wang, F.; Fu, X.; Huang, J.~Y.; Li, Z.; Liu, Q.; Liu, X.; Ma, M.~D.; Xu, N.; Zhou, W.; Zhang, K.; et~al. 2024{\natexlab{a}}.
\newblock Muirbench: A comprehensive benchmark for robust multi-image understanding.
\newblock \emph{arXiv preprint arXiv:2406.09411}.

\bibitem[{Wang et~al.(2025)Wang, Qu, Huang, Chu, Lin, and Chen}]{wang2025vl-rethinker}
Wang, H.; Qu, C.; Huang, Z.; Chu, W.; Lin, F.; and Chen, W. 2025.
\newblock Vl-rethinker: Incentivizing self-reflection of vision-language models with reinforcement learning.
\newblock \emph{arXiv preprint arXiv:2504.08837}.

\bibitem[{Wang and Ke(2024)}]{wang2024llmseg}
Wang, J.; and Ke, L. 2024.
\newblock Llm-seg: Bridging image segmentation and large language model reasoning.
\newblock In \emph{Proceedings of the IEEE/CVF Conference on Computer Vision and Pattern Recognition}, 1765--1774.

\bibitem[{Wang et~al.(2024{\natexlab{b}})Wang, Bai, Tan, Wang, Fan, Bai, Chen, Liu, Wang, Ge et~al.}]{wang2024qwen2vl}
Wang, P.; Bai, S.; Tan, S.; Wang, S.; Fan, Z.; Bai, J.; Chen, K.; Liu, X.; Wang, J.; Ge, W.; et~al. 2024{\natexlab{b}}.
\newblock Qwen2-vl: Enhancing vision-language model's perception of the world at any resolution.
\newblock \emph{arXiv preprint arXiv:2409.12191}.

\bibitem[{Wang et~al.(2024{\natexlab{c}})Wang, Lv, Yu, Hong, Qi, Wang, Ji, Yang, Zhao, XiXuan et~al.}]{wang2024cogvlm}
Wang, W.; Lv, Q.; Yu, W.; Hong, W.; Qi, J.; Wang, Y.; Ji, J.; Yang, Z.; Zhao, L.; XiXuan, S.; et~al. 2024{\natexlab{c}}.
\newblock Cogvlm: Visual expert for pretrained language models.
\newblock \emph{Advances in Neural Information Processing Systems}, 37: 121475--121499.

\bibitem[{Wu et~al.(2024)Wu, Yu, Chen, Tenenbaum, and Gan}]{wu2024star}
Wu, B.; Yu, S.; Chen, Z.; Tenenbaum, J.~B.; and Gan, C. 2024.
\newblock Star: A benchmark for situated reasoning in real-world videos.
\newblock \emph{arXiv preprint arXiv:2405.09711}.

\bibitem[{Xie et~al.(2023)Xie, Zhang, Wu, Zhu, Zhao, and Liang}]{xie2023d3}
Xie, C.; Zhang, Z.; Wu, Y.; Zhu, F.; Zhao, R.; and Liang, S. 2023.
\newblock Described Object Detection: Liberating Object Detection with Flexible Expressions.
\newblock In \emph{Thirty-seventh Conference on Neural Information Processing Systems (NeurIPS)}.

\bibitem[{Xu et~al.(2025)Xu, Zhu, Jiang, Li, Shen, Wang, Huang, Wang, Zhang, Yang et~al.}]{xu2025mitracker}
Xu, M.; Zhu, Y.; Jiang, H.; Li, J.; Shen, Z.; Wang, S.; Huang, H.; Wang, X.; Zhang, H.; Yang, Q.; et~al. 2025.
\newblock MITracker: Multi-View Integration for Visual Object Tracking.
\newblock In \emph{Proceedings of the Computer Vision and Pattern Recognition Conference}, 27176--27185.

\bibitem[{Xu, Zhu, and Yang(2024)}]{xu2024mc-bench}
Xu, Y.; Zhu, L.; and Yang, Y. 2024.
\newblock Mc-bench: A benchmark for multi-context visual grounding in the era of mllms.
\newblock \emph{arXiv preprint arXiv:2410.12332}.

\bibitem[{Yan et~al.(2024)Yan, Wang, Yan, Jiang, Hu, Kang, Xie, and Gavves}]{yan2024revos}
Yan, C.; Wang, H.; Yan, S.; Jiang, X.; Hu, Y.; Kang, G.; Xie, W.; and Gavves, E. 2024.
\newblock Visa: Reasoning video object segmentation via large language models.
\newblock In \emph{European Conference on Computer Vision}, 98--115. Springer.

\bibitem[{Yang et~al.(2025)Yang, He, Pan, Jiang, Deng, Yang, Lu, Yin, Rao, Zhu et~al.}]{yang2025r1-onevision}
Yang, Y.; He, X.; Pan, H.; Jiang, X.; Deng, Y.; Yang, X.; Lu, H.; Yin, D.; Rao, F.; Zhu, M.; et~al. 2025.
\newblock R1-onevision: Advancing generalized multimodal reasoning through cross-modal formalization.
\newblock \emph{arXiv preprint arXiv:2503.10615}.

\bibitem[{Yao et~al.(2024)Yao, Yu, Zhang, Wang, Cui, Zhu, Cai, Li, Zhao, He et~al.}]{yao2024minicpm-v}
Yao, Y.; Yu, T.; Zhang, A.; Wang, C.; Cui, J.; Zhu, H.; Cai, T.; Li, H.; Zhao, W.; He, Z.; et~al. 2024.
\newblock Minicpm-v: A gpt-4v level mllm on your phone.
\newblock \emph{arXiv preprint arXiv:2408.01800}.

\bibitem[{Ye et~al.(2024)Ye, Xu, Liu, Hu, Yan, Qian, Zhang, Huang, and Zhou}]{ye2024mplug-owl3}
Ye, J.; Xu, H.; Liu, H.; Hu, A.; Yan, M.; Qian, Q.; Zhang, J.; Huang, F.; and Zhou, J. 2024.
\newblock mplug-owl3: Towards long image-sequence understanding in multi-modal large language models.
\newblock \emph{arXiv preprint arXiv:2408.04840}.

\bibitem[{You et~al.(2023)You, Zhang, Gan, Du, Zhang, Wang, Cao, Chang, and Yang}]{you2023ferret}
You, H.; Zhang, H.; Gan, Z.; Du, X.; Zhang, B.; Wang, Z.; Cao, L.; Chang, S.-F.; and Yang, Y. 2023.
\newblock Ferret: Refer and ground anything anywhere at any granularity.
\newblock \emph{arXiv preprint arXiv:2310.07704}.

\bibitem[{Yu et~al.(2025)Yu, Lin, Zhao, Yin, Wei, Peng, Wei, Sun, Han, Ge et~al.}]{yu2025perception-r1}
Yu, E.; Lin, K.; Zhao, L.; Yin, J.; Wei, Y.; Peng, Y.; Wei, H.; Sun, J.; Han, C.; Ge, Z.; et~al. 2025.
\newblock Perception-r1: Pioneering perception policy with reinforcement learning.
\newblock \emph{arXiv preprint arXiv:2504.07954}.

\bibitem[{Yu et~al.(2016)Yu, Poirson, Yang, Berg, and Berg}]{yu2016refcoco}
Yu, L.; Poirson, P.; Yang, S.; Berg, A.~C.; and Berg, T.~L. 2016.
\newblock Modeling context in referring expressions.
\newblock In \emph{European conference on computer vision}, 69--85. Springer.

\bibitem[{Zhan et~al.(2024{\natexlab{a}})Zhan, Zhu, Chen, Yang, Tang, and Wang}]{zhan2024griffon}
Zhan, Y.; Zhu, Y.; Chen, Z.; Yang, F.; Tang, M.; and Wang, J. 2024{\natexlab{a}}.
\newblock Griffon: Spelling out all object locations at any granularity with large language models.
\newblock In \emph{European Conference on Computer Vision}, 405--422. Springer.

\bibitem[{Zhan et~al.(2024{\natexlab{b}})Zhan, Zhu, Zhao, Yang, Tang, and Wang}]{zhan2024griffonv2}
Zhan, Y.; Zhu, Y.; Zhao, H.; Yang, F.; Tang, M.; and Wang, J. 2024{\natexlab{b}}.
\newblock Griffon v2: Advancing multimodal perception with high-resolution scaling and visual-language co-referring.
\newblock \emph{arXiv preprint arXiv:2403.09333}.

\bibitem[{Zhan et~al.(2025)Zhan, Zhu, Zheng, Zhao, Yang, Tang, and Wang}]{zhan2025vision-r1}
Zhan, Y.; Zhu, Y.; Zheng, S.; Zhao, H.; Yang, F.; Tang, M.; and Wang, J. 2025.
\newblock Vision-r1: Evolving human-free alignment in large vision-language models via vision-guided reinforcement learning.
\newblock \emph{arXiv preprint arXiv:2503.18013}.

\bibitem[{Zhang et~al.(2025)Zhang, Huang, Yao, Liu, Zhang, Lu, and Tao}]{zhang2025r1-vl}
Zhang, J.; Huang, J.; Yao, H.; Liu, S.; Zhang, X.; Lu, S.; and Tao, D. 2025.
\newblock R1-vl: Learning to reason with multimodal large language models via step-wise group relative policy optimization.
\newblock \emph{arXiv preprint arXiv:2503.12937}.

\end{thebibliography}
}
\maketitlesupplementary

\section{Dataset Details}
As demonstrated in the Data Construction of the main paper, we categorize multi-image grounding tasks and extend existing datasets~\cite{li2025migician} by addressing the limitations in instance quantities and cross-image relations. We collect and process 240K samples to build our dataset. In this section, we detail the task taxonomy, data construction and training data as below.

\subsection{Task Taxonomy}

As illustrated in Figure~2(b) in the main paper, we categorize multi-image grounding tasks into three main types based on their reliance on cross-image cues and reasoning: referring retrieval grounding, cross-image association grounding, and reasoning grounding.

\begin{itemize}
    \item Referring retrieval grounding does not require cross-image comparison or reasoning. Instead, the instructions identify instances using their categories, attributes, locations, or states explicitly. While no cross-image comparison is needed, the presence of distractors from negative images increases the task’s complexity, making it more challenging than single-image grounding.
    \item Cross-image association grounding involves identifying targets through across-image visual correspondences rather than direct descriptions. According to the image relations, it can be further divided into: Semantic association grounding, which focuses on attributes, categories, quantities, and interactions (e.g., identifying the common objects); temporal association grounding, where models need to perceive temporal cues across sequential frames; spatial association grounding, where models are required to understand scenes from varying viewpoints.
    \item Reasoning grounding, unlike the perception-focused tasks above, requires models to use commonsense or external knowledge to locate instances.

\end{itemize}

\subsection{Data Construction}
\begin{table}[t]
\centering
\resizebox{1.0\columnwidth}{!}{
\begin{tabular}{c|cc}
\toprule
\textbf{Type} & \textbf{Source} & \textbf{Vol.} \\ 
\midrule 
\multicolumn{3}{c}{\textbf{Stage-1 (SFT)}} \\ 

\midrule 
\multirow{1}{*}{M-Understanding} & M4-Instruct~\cite{li2024llava-next-interleave} & 200K \\ 
\multirow{2}{*}{M-Grounding} & MGrounding-630k~\cite{li2025migician} & 630K \\ 
& self-construct & 240K \\
\midrule 
 
\multicolumn{3}{c}{\textbf{Stage-2 (SFT)}} \\ 
\midrule  
\multirow{3}{*}{S-Grounding} & GRefCOCO~\cite{grefcoco} & 30K \\ 
& ODINW~\cite{li2022odinw} & 30K \\ 
& RefCOCO~\cite{yu2016refcoco} & 14K \\ 
\multirow{2}{*}{M-Grounding}  & UniVG-R1-SFTData~\cite{bai2025univg-r1} & 76K \\ 
 & self-construct & 158K \\ 
 \midrule 

\multicolumn{3}{c}{\textbf{Stage-3 (RL)}} \\ 
\midrule  
\multirow{2}{*}{S-Grounding} & RefCOCO~\cite{yu2016refcoco} & 3K \\ 
& ODINW~\cite{li2022odinw} & 2K \\  
\multirow{2}{*}{M-Grounding}  & UniVG-R1-RLData~\cite{bai2025univg-r1} & 7K \\ 
 & self-construct & 14K \\ 
 
\bottomrule
\end{tabular}%
}
\caption{Training data for each stage.}
\label{tab:train data detail}
\end{table}

\begin{figure*}[ht!]
	\centering
	\begin{minipage}[t]{\linewidth}
		\centering
		\includegraphics[width=0.98\linewidth]{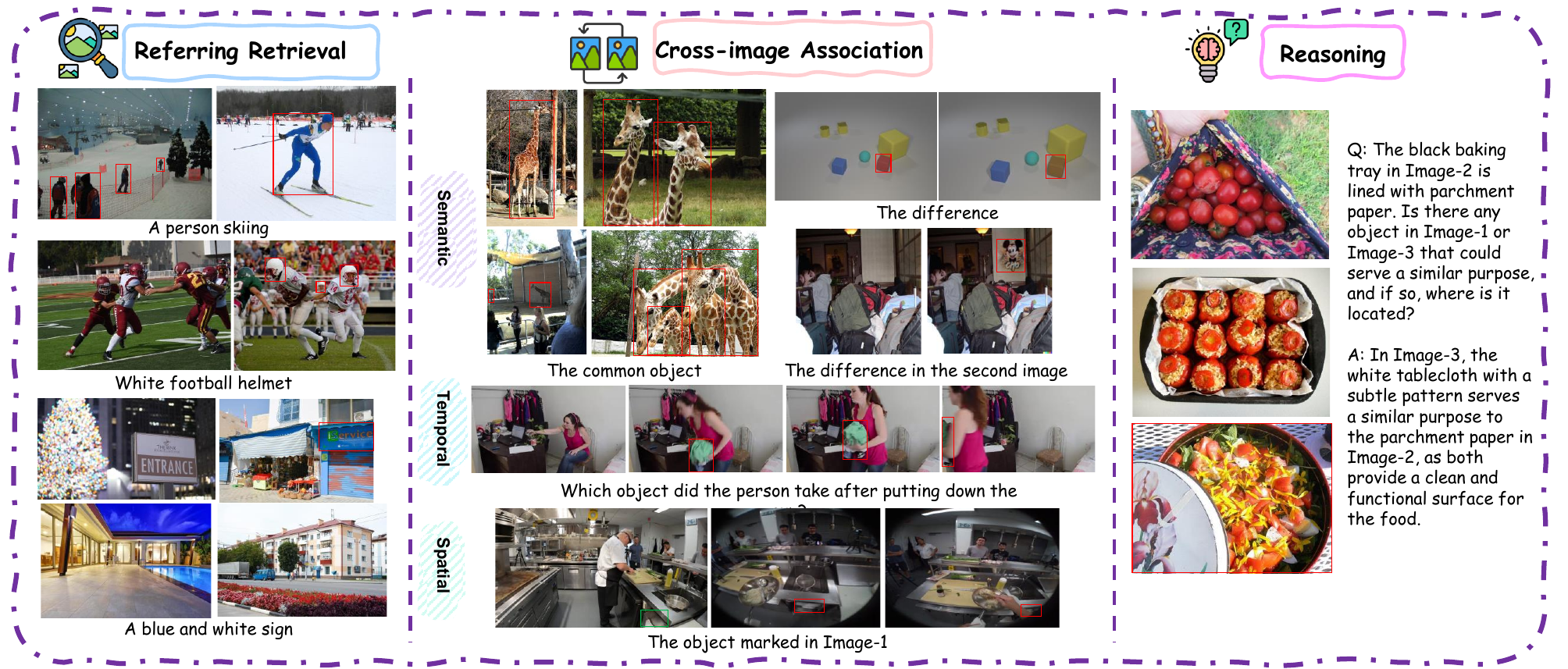}
		\caption{Examples of training data within each type of tasks. For reasoning grounding task, we use the free-form grounding data from MGrounding-630K.} 
        \label{fig:train data examples}
	\end{minipage}
\end{figure*}

In this subsection, we detail the data construction pipeline for each task type.
\subsubsection{Referring retrieval grounding}
The data for this category is sourced from the $D^3$~\cite{xie2023d3} and COCO~\cite{lin2014mscoco} datasets. $D^3$ provides complete group-wise annotations of target objects described by given expressions. We first exclude the images that appeared in MC-Bench from the original dataset. Since the source images are already grouped by scene, we directly extract all expressions and sample image pairs within the corresponding image pool of each expression. During sampling, the selection probabilities of positive and negative images are both set to 0.5 to ensure a diverse target distribution within the training samples.
For the COCO dataset, we first collect a pool of images for each category based on annotated labels and filter out those with 2 to 10 bounding boxes of the corresponding category. We then randomly sample 2 to 4 images from each category pool to form an image group, with some of the sampled images intentionally selected as negative examples from other category pools.
\subsubsection{Semantic association grounding}
For cross-image semantic associations, we focus on locating objects of the same category across images. Unlike MGrounding-630K, which targets a single primary common object in each image, our setting requires identifying all instances of the shared category to promote multi-object localization. The data for this task is based on the COCO dataset. Specifically, we first organize image pools for each category and reserve images containing 1 to 3 bounding boxes of the target category. We then randomly group 2 to 4 images from the same pool. After forming a group, we check for any additional shared categories among the selected images and include them in the grounding targets as well.
\subsubsection{Spatial association grounding}
We select Ego-Exo4D~\cite{grauman2024ego-exo4d} and MVTrack~\cite{xu2025mitracker} as our data sources, as they contain a rich set of multi-view images compared to traditional object tracking datasets. Ego-Exo4D centers around simultaneously captured egocentric and exocentric videos of human activities. Based on the relation annotations, we assign a balanced number of image groups to each object within a take to ensure diverse scene and object coverage. Each group consists of 2 to 4 frames sampled from egocentric and exocentric views, with considerable variations in target locations. This design creates more challenging training samples, requiring the model to identify the same object across noticeably different viewpoints.
MVTrack provides annotations for a single object per take, with fixed camera viewpoints and counts. For each take, we randomly sample a number of image groups, where each group contains four images captured from different views with the same timestamp.

\subsubsection{Temporal association grounding}
We utilize the STAR~\cite{wu2024star} dataset, a situation reasoning dataset built upon real-world videos associated with human actions and interactions. Since many of the original questions focus on action inference, we reformulate them into object-centric queries and extract object location information from the situation graphs. To construct image groups, we retain the first and last frames of each original sequence and randomly sample 2 to 4 additional frames from the middle segment that contain annotations for the target object.

\subsection{Details of Training Data}
This subsection outlines the data sources and statistics for three training stages, as summarized in Table~\ref{tab:train data detail}.

In Stage 1, we employ the multi-image dataset encompassing grounding and general understanding tasks to enhance the model’s foundational capabilities in multi-image localization and comprehension. A substantial portion of the training data is sourced from MGrounding-630K and our self-constructed multi-image grounding dataset, as our primary focus is on multi-image grounding and the base model already demonstrates strong general multi-image understanding capabilities. Figure~\ref{fig:train data examples} illustrates a few examples of the multi-image grounding data.

In Stage 2, we primarily use the UniVG-R1 supervised finetuning data and our self-constructed dataset to teach the model the reasoning forms required by multi-image grounding tasks and maintain the previously learned capabilities. Additionally, single-image grounding data is incorporated to preserve the base model’s performance on single-image localization. Similar to the multi-image setting, both single-target and multi-target scenarios are considered for single-image grounding data.

In Stage 3, the reinforcement learning stage, part of the multi-image grounding data are sourced from UniVG-R1 RL data, while the remaining are sampled from prior stages.

\begin{table*}[t]
    \centering
    \resizebox{0.95\linewidth}{!}{
    \begin{tabular}{lcccccccccccccc}
        \toprule
         \multirow{1}{*}[2.0 ex]{Models} & 
           \tiny \rotatebox{75}{AerialDrone} & \tiny \rotatebox{75}{Aquarium} & \tiny \rotatebox{75}{Rabbits} & \tiny \rotatebox{75}{EgoHands} & \tiny \rotatebox{75}{Mushrooms} & \tiny \rotatebox{75}{Packages} & \tiny \rotatebox{75}{PascalVOC} & \tiny \rotatebox{75}{pistols} & \tiny \rotatebox{75}{pothole} & \tiny \rotatebox{75}{Raccoon} & \tiny \rotatebox{75}{Shellfish} & \tiny \rotatebox{75}{thermal} & \tiny \rotatebox{75}{Vehicles} & \multirow{1}{*}[2.0 ex]{AVG}\\
         \midrule
         MDETR~\cite{kamath2021mdetr}  & 0.6 & 1.7 & 66.5 & 5.9 & 39.8 & 63.6 & 5.6 & 15.9 & 12.7 & 50.6 & 8.1 & 4.5 & 13.4 & 22.2  \\
         G-DINO-L~\cite{liu2024groundingdino} & 12.6 & 28.1 & 71.7 & 52.0 & 72.3 & 63.9 & 66.0 & 71.4 & 30.4 & 65.8 & 62.5 & 21.3 & 62.7 & 52.4 \\
         \midrule
         Ferret-13B\dag\;~\cite{you2023ferret} & 0 & 4.3 & 59.8 & 1.5 & 6.1 & 40.1 & 35.2 & 41.5 & 3.9 & 49.5 & 29.5 & 36.5 & 44.4 & 27.1 \\
         InternVL2.5-8B\dag\;~\cite{chen2024internvl2.5} & 0 & 6.9 & 38.5 & 0.2 & 26.7 & 16.4 & 37.0 & 29.2 & 1.1 & 46.6 & 28.5 & 3.8 & 27.1 & 20.2 \\
         Qwen2-VL-7B\dag\;~\cite{wang2024qwen2vl} & 4.9 & 14.2 & 65.8 & 11.8 & 3.7 & 47.5 & 44.4 & 52.1 & \textbf{12.1} & 40.3 & 38.9 & 32.2 & 48.6 & 32.0 \\
         Migician\dag\;~\cite{li2025migician} & 0.9 & 4.3 & 54.5 & 2.9 & 17.6 & 15.1 & 28.7 & 36.7 & 5.0 & 33.3 & 28.9 & 23.0 & 33.6 & 21.9 \\
         UniVG-R1\dag\;~\cite{bai2025univg-r1} & 0.6 & 10.5 & 66.5 & 1.1 & 30.8 & 52.5 & 44.2 & 50.7 & 10.9 & 40.5 & 36.7 & 37.2 & 51.0 & 33.3 \\
         Qwen2.5-VL-3B ~\cite{bai2025qwen25vl}& 6.2 & 16.4 & 75.0 & 24.6 & 8.3 & 66.6 & 52.0 & 42.3 & 10.2 & 47.7 & 36.7 & 40.7 & 57.1 & 37.2 \\
         Qwen2.5-VL-7B\dag\;~\cite{bai2025qwen25vl} & 7.8 & 20.3 & 73.5 & 32.2 & 7.0 & 57.6 & 49.8 & 48.5 & 7.4 & 40.1 & 42.7 & 38.0 & 56.3 & 37.0 \\
         \midrule
         GeM-VG & 3.4 & 25.6 & 77.4 & 51.0 & 20.6 & 51.0 & 54.1 & 57.5 & 12.1 & 36.2 & 40.6 & 44.3 & 60.3 & 41.1 \\
         \bottomrule
    \end{tabular}}
    \caption{Detailed results on ODINW-13 dataset. \dag\; indicates the results are reproduced to get results of all sets following the official Settings.}
    \label{tab:ODINW results}
\end{table*}

\begin{table*}[t]
    \centering
    \resizebox{0.95\linewidth}{!}{
    \begin{tabular}{lccccccccccccc}
        \toprule
         \multirow{1}{*}[2.0 ex]{Models} & \multirow{1}{*}[2.0 ex]{Overall} &
           \tiny \rotatebox{75}{Action.} & \tiny \rotatebox{75}{Attribute.} & \tiny \rotatebox{75}{cartoon.} & \tiny \rotatebox{75}{Counting} & \tiny \rotatebox{75}{Diagram.} & \tiny \rotatebox{75}{Difference.} & \tiny \rotatebox{75}{Geographic.} & \tiny \rotatebox{75}{Matching.} & \tiny \rotatebox{75}{Ordering} & \tiny \rotatebox{75}{Scene.} & \tiny \rotatebox{75}{Grounding.} & \tiny \rotatebox{75}{Retrieval.} \\
         \midrule
         GPT-4o~\cite{openai2023gpt4o} & 68.00 & 44.51 & 56.12 & 51.28 & 49.15 & 88.69 & 60.29 & 56.00 & 86.85 & 23.44 & 71.51 & 36.90 & 80.14  \\
         Gemini Pro~\cite{team2024geminiPRO} & 49.35 & 35.98 & 41.33 & 47.44 & 28.63 & 64.82 & 45.29 & 48.00 & 66.59 & 12.50 & 59.14 & 28.57 & 43.84  \\
         \midrule
         LLaVA-1.5~\cite{liu2024llava1.5} & 23.46 & 27.44 & 22.96 & 24.36 & 23.08 & 25.13 & 20.00 & 20.00 & 23.49 & 23.44 & 34.95 & 14.29 & 19.86  \\
         CogVLM~\cite{wang2024cogvlm} & 20.85 & 26.22 & 16.33 & 41.03 & 14.10 & 19.60 & 19.71 & 13.00 & 21.34 & 12.50 & 41.40 & 16.67 & 15.75 \\
         Idefics2-8B~\cite{laurenccon2024idefics2} & 26.08 & 26.22 & 17.86 & 39.74 & 21.79 & 25.38 & 27.65 & 21.00 & 24.78 & 15.62 & 56.45 & 26.19 & 17.12  \\
         Mantis~\cite{jiang2024mantis} & 44.50 & 33.54 & 48.47 & 38.46 & 38.46 & 67.59 & 28.82 & 26.00 & 53.88 & 18.75 & 56.99 & 26.19 & 35.62  \\
         Qwen2-VL-7B~\cite{wang2024qwen2vl} & 39.88 & 38.41 & 49.49 & 42.31 & 40.60 & 39.70 & 31.76 & 26.00 & 51.51 & 12.50 & 68.28 & 32.14 & 19.18  \\
         Migician~\cite{li2025migician} & 57.81 & 54.27 & 50.00 & 58.97 & 55.13 & 65.83 & 52.65 & 50.00 & 63.58 & 50.00 & 74.19 & 46.43 & 50.00  \\
         UniVG-R1~\cite{bai2025univg-r1} & 44.77 & 39.63 & 46.94 & 44.87 & 42.74 & 52.26 & 44.12 & 27.00 & 55.17 & 12.50 & 67.74 & 30.95 & 24.32  \\
         \midrule
         GeM-VG & 58.20 & 54.88 & 50.00 & 64.10 & 60.68 & 68.34 & 57.35 & 50.00 & 59.05 & 50.00 & 66.13 & 48.81 & 50.00  \\
        
         \bottomrule
    \end{tabular}}
    \caption{Results on MuirBench.}
    \label{tab:MuirBench results}
\end{table*}

\section{Detailed Results}
Due to the page length limitation, we do not present the detailed results of some benchmarks in the main body. In this section, we provide the detailed results on ODINW-13~\cite{li2022odinw} and MuirBench~\cite{wang2024muirbench} benchmarks, to clearly demonstrate GeM-VG's superior capabilities in multi-target grounding and general multi-image understanding.

ODINW-13 is an object detection dataset with rare categories and various numbers of targets in the wild. It consists of 13 distinct subsets, evaluating the model's capability to identify and ground objects in diverse practical scenarios.
As shown in Table~\ref{tab:ODINW results}, our model outperforms other models across multiple sets, especially when there are multiple queried objects, such as Aquarium, Egohands and PascalVOC. Compared to the base model Qwen2-VL-7B~\cite{wang2024qwen2vl}, our model achieves an average improvement of 9.1\%, showcasing the superiority of our model in multiple object grounding. The recent Migician~\cite{li2025migician} and UniVG-R1~\cite{bai2025univg-r1}, in contrast, face challenges in such intensive object grounding, as their training data and strategy are limited to single-target scenarios.

MuirBench~\cite{wang2024muirbench} is a comprehensive multi-image understanding benchmark, consisting of 12 diverse multi-image tasks. We list the results of all subtasks from different models in Table~\ref{tab:MuirBench results} for further analysis. 
Our model achieves notable improvements on tasks such as Counting, Difference Spotting, and Grounding, which require fine-grained visual perception and generalized multi-image grounding capabilities. Specifically, GeM-VG outperforms Migician by 5.55\% on Counting and 2.38\% on Grounding. While another multi-image grounding model, UniVG-R1, lags behind Migician, our model surpasses it in overall score, demonstrating robust general multi-image understanding capabilities.

\section{More Ablations}
\subsubsection{Hyperparameter Setting}
During RL training with a mixed fine-tuning strategy, to prevent the response patterns from prematurely collapsing into a single mode in the early stage, we modulate the reward according to the proportion of different response modes within each batch, as shown in Equation~\ref{eq:adjust_early}. The parameter $p$ is empirically set to $0.5$ to balance the output probabilities of the two response modes during the early phase of RL training.
For $\alpha$, which controls the magnitude of reward modulation, we evaluate values of $0.5$, $2.0$, and $3.0$ while keeping all other training settings fixed. The corresponding results on MIG-Bench are reported in Table~\ref{tab:hyp}. When $\alpha$ is too small, the modulation is insufficient, leading to an imbalance between the two response modes in the early stage. Conversely, an excessively large $\alpha$ disrupts the original reward distribution, preventing the model from optimizing toward more accurate responses.
For $\delta_1$ and $\delta_2$, we set $\delta_2 < \delta_1$ to prevent inaccurate rollouts from dominating as much as possible, thereby stabilizing the proportion adjustment while minimizing interference with the relative ordering of the original rewards.

\begin{table}[t!]
\centering
\resizebox{0.6\linewidth}{!}{
\begin{tabular}{l|c|c|c}
\hline
\diagbox{$\delta_1 / \delta_2$}{$\alpha$} & 0.5 & 2.0 & 3.0 \\
\hline
1.0 / 0.5 & 74.30 & 74.96 & 73.85  \\
\hline
1.0 / 1.0 & - & 73.66 & -  \\
\hline
\end{tabular}
}
\caption{Ablation results on MIG-Bench with different reward modulation hyperparameters.}
\label{tab:hyp}
\end{table}

\subsubsection{Effectiveness of MIG-Data-240K}
The MIG-Data-240K dataset comprises diverse multi-image localization task types, with the majority involving multi-instance targets, and is designed to train the model’s capability for multi-image, multi-object localization. When MIG-Data-240K is removed from the training set, the AP@50 on MC-Bench drops to 22.4, and the average score on MIG-Bench decreases to 72.04, corresponding to relative reductions of 10.4\% and 2.61\%, respectively.
\subsubsection{Results on other Model}
To verify the applicability of our method to other MLLMs, we conduct an experiment on Qwen2.5-VL-7B using the same Stage-2 (SFT) and Stage-3 (RL) training data and identical training strategies. The resulting model achieves scores of 69.68 on MIG-Bench and 39.7 on MC-Bench, respectively, demonstrating that our approach generalizes well to other MLLMs.

\section{Qualitative Analysis}

\begin{figure*}[ht!]
	\centering
	\begin{minipage}[t]{\linewidth}
		\centering
		\includegraphics[width=0.95\linewidth]{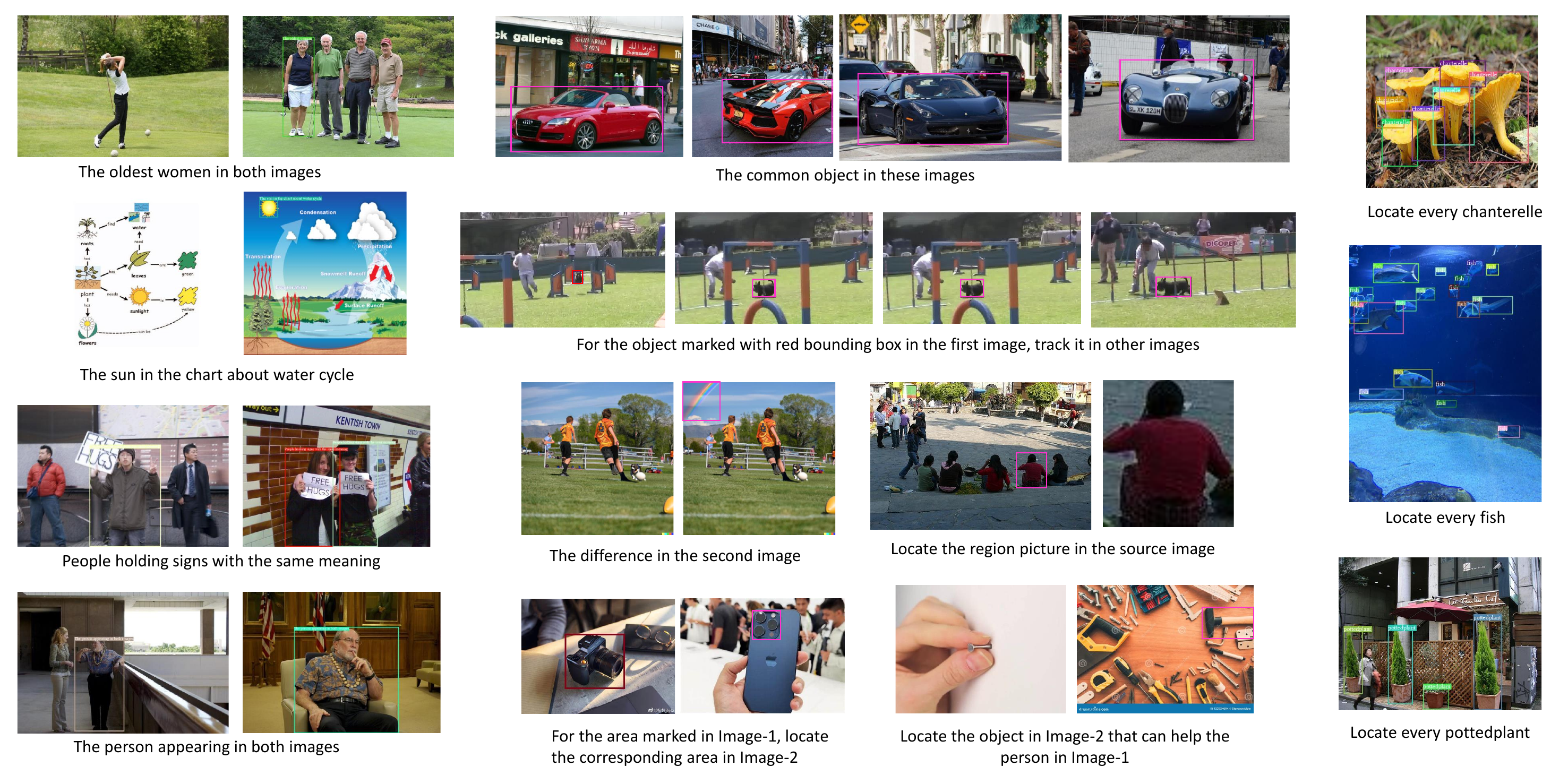}
		\caption{Some visualization examples on grounding task. GeM-VG is capable of a wide range of grounding tasks.} 
        \label{fig:vis all}
	\end{minipage}
\end{figure*}

\begin{figure*}[ht!]
	\centering
	\begin{minipage}[t]{\linewidth}
		\centering
		\includegraphics[width=0.95\linewidth]{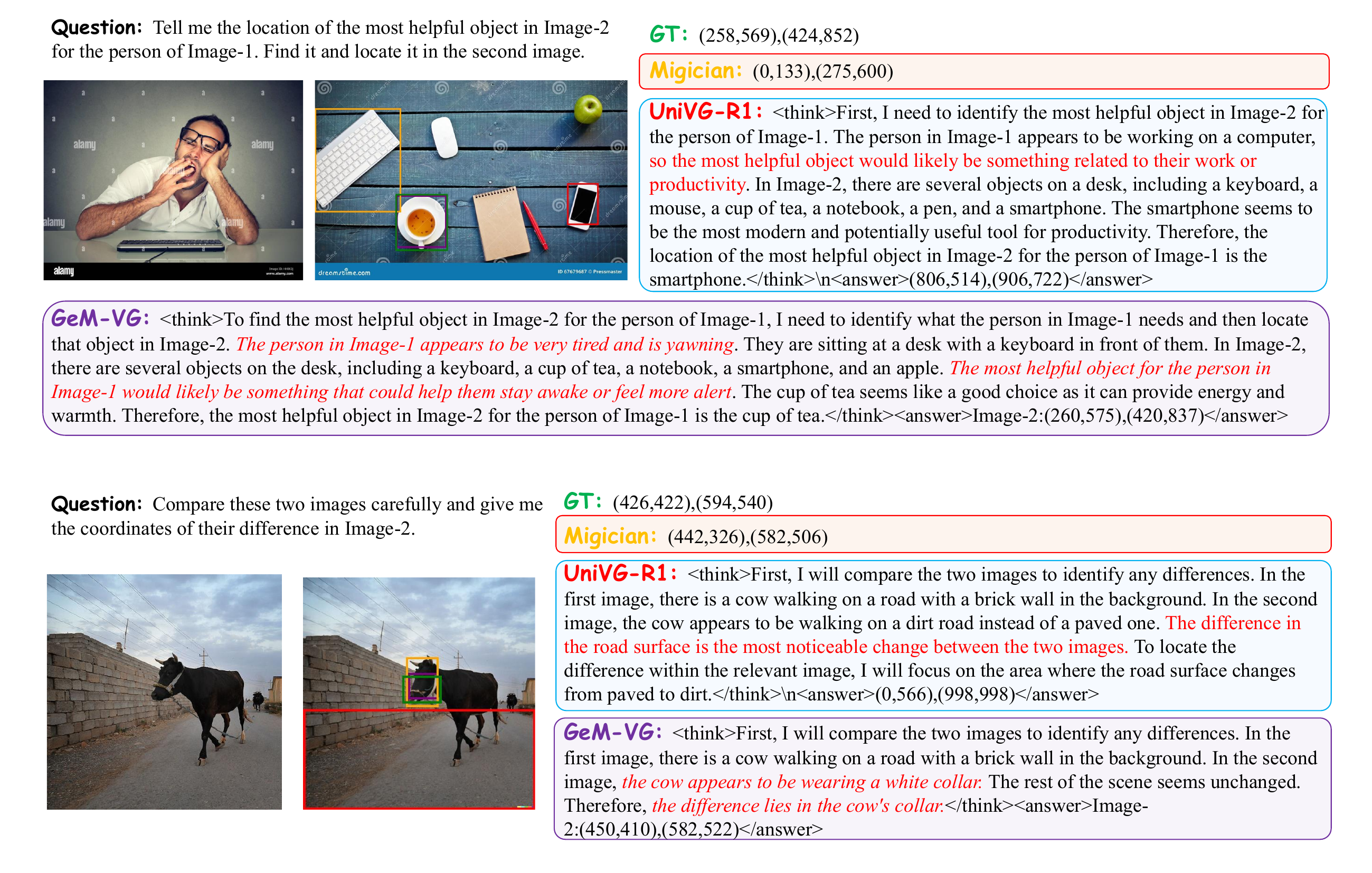}
		\caption{Qualitative comparison. } 
        \label{fig:vis compare}
	\end{minipage}
\end{figure*}

Our model can handle a wide range of visual grounding tasks. We present the visualization results on different tasks in Figure~\ref{fig:vis all}. To better demonstrate the grounding and reasoning capabilities of our model, we provide a qualitative comparison with Migician and UniVG-R1 in Figure~\ref{fig:vis compare}.
While Migician struggles with grounding tasks that require reasoning (e.g., determining which object can be helpful to the person in the top case) and UniVG-R1 hallucinates in tasks demanding fine-grained semantic discrimination (e.g., identifying subtle differences between images in the bottom case), GeM-VG consistently provides more accurate results and explanations across both perception-focused and reasoning-focused tasks. These comparison results clearly demonstrate our model’s enhanced capabilities in multi-image perception and reasoning.

\end{document}